\documentclass{article} 
\usepackage[final]{colm2025_conference}

\usepackage{changepage}
\usepackage{microtype}
\usepackage{hyperref}
\usepackage{url}
\usepackage{booktabs}
\usepackage{wrapfig} 
\usepackage{graphicx}
\usepackage{caption}
\usepackage{xspace}
\usepackage{multirow}
\usepackage{fancyvrb}
\usepackage{fvextra}
\usepackage{fontawesome}
\usepackage[most,breakable]{tcolorbox}
\usepackage{enumitem}
\usepackage{pifont}
\usepackage{amssymb}

\usepackage{multirow}
\usepackage{diagbox}
\usepackage{makecell}
\usepackage{colortbl}
\usepackage{mdframed}

\usepackage{pgfplots}
\usepackage{pgfplotstable}
\pgfplotsset{compat=1.18}

\usepackage{siunitx}

\definecolor{darkblue}{rgb}{0, 0, 0.5}
\hypersetup{colorlinks=true, citecolor=darkblue, linkcolor=darkblue, urlcolor=darkblue}

\newcommand{\finemath}{MegaMath\xspace}

\newcommand{\owm}{{Open-Web-Math}\xspace}
\newcommand{\mathpile}{{MathPile}\xspace}
\newcommand{\infimm}{{Infimm-Web-Math}\xspace}

\newcommand{\dsmath}{DeepSeekMath\xspace}

\newcommand{\llamaiii}{{Llama-3}\xspace}

\newcommand{\resili}{\texttt{Resiliparse}\xspace}
\newcommand{\trafil}{\texttt{trafilatura}\xspace}
\newcommand{\fasttext}{\texttt{fastText}}
\newcommand{\huggingface}{\raisebox{-1.5pt}{\includegraphics[height=1.05em]{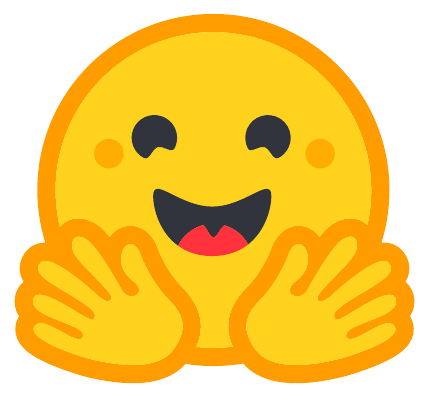}}\xspace}
\newcommand{\github}{\raisebox{-1.5pt}{\includegraphics[height=1.05em]{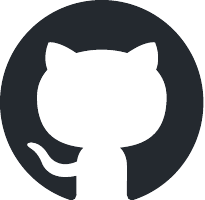}}\xspace}

\newmdenv[
  backgroundcolor=red!05,
  linecolor=quoteborder,
  skipabove=1em,
  skipbelow=0em,
  leftline=true,
  topline=false,
  bottomline=false,
  rightline=false,
  linecolor=red!66,
  linewidth=4pt
]{lightquote}

\title{
\finemath: Pushing the Limits of Open Math Corpora
}

\author{Fan Zhou\thanks{Equal Contribution.} \quad Zengzhi Wang$^{*}$ \quad Nikhil Ranjan \quad 
Zhoujun Cheng \quad
Liping Tang \\ 
\textbf{Guowei He \quad Zhengzhong Liu \quad Eric P. Xing} \\
{MBZUAI}\\
\\
\huggingface \url{{https://hf.co/datasets/LLM360/MegaMath}}\\
\github \url{https://github.com/LLM360/MegaMath}\\
}

\begin{document}

\ifcolmsubmission
\linenumbers
\fi

\maketitle

\begin{abstract}
Mathematical reasoning is a cornerstone of human intelligence and a key benchmark for advanced capabilities in large language models (LLMs).
However, the research community still lacks an open, large-scale, high-quality corpus tailored to the demands of math-centric LLM pre-training.
We present \finemath, an open dataset curated from diverse, math-focused sources through following practices:
(1) \textbf{\textit{Revisiting web data}}:
We re-extracted mathematical documents from Common Crawl with math-oriented HTML optimizations, \fasttext-based filtering and deduplication, all for acquiring higher-quality data on the Internet.
(2) \textbf{\textit{Recalling Math-related code data}}: We identified high quality math-related code from large code training corpus, Stack-V2, further enhancing data diversity.
(3) \textbf{\textit{Exploring Synthetic data}}: We synthesized QA-style text, math-related code, and interleaved text-code blocks from web data or code data. By integrating these strategies and validating their effectiveness through extensive ablations, \finemath delivers \textbf{371B tokens} with the largest quantity and top quality among existing open math pre-training datasets.

\end{abstract}

\begin{figure}[htbp]
\vspace{-1.5mm}
    \centering
    \includegraphics[width=0.9\textwidth]{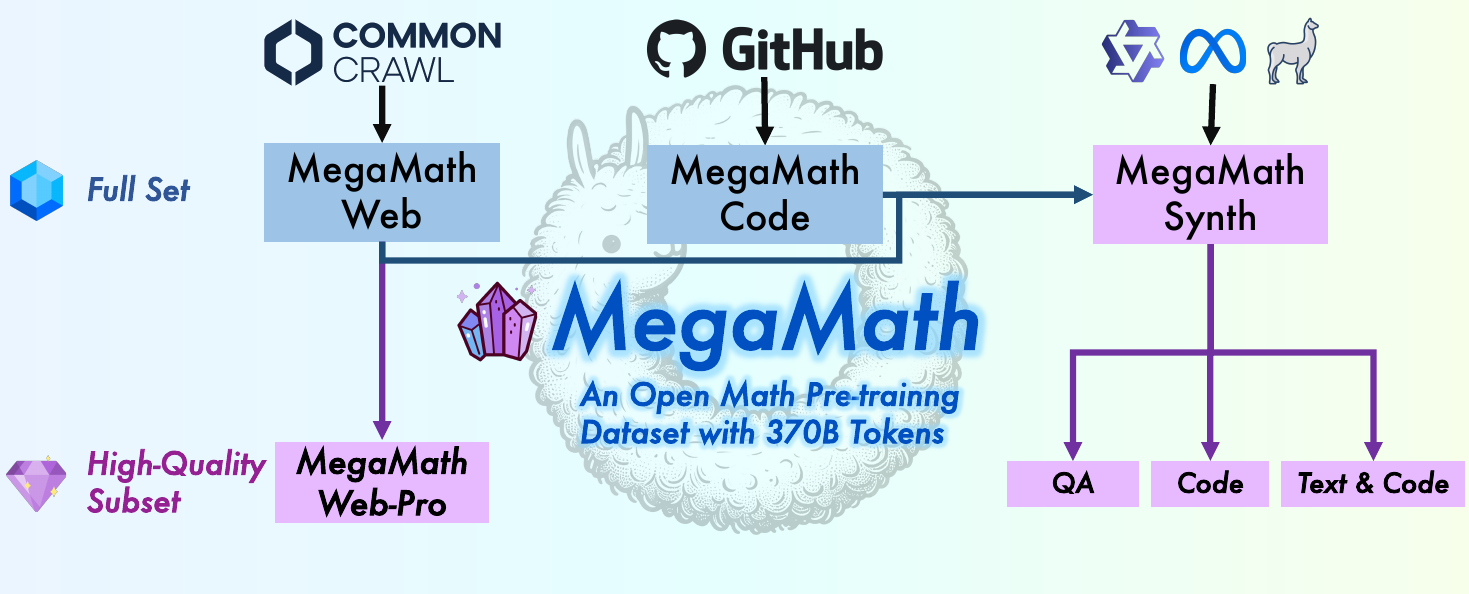}
    \caption{
    The overview of \finemath dataset.
    }
    \label{fig:data_overview}
\end{figure}

\section{Introduction}

Mathematical reasoning is a fundamental yet challenging aspect of human intelligence—and a persistent difficulty for language models. Recent breakthroughs like o1~\citep{openaio1} and DeepSeek-R1~\citep{guo2025deepseekr1} demonstrate that, with sufficient pre-training and large-scale reinforcement learning, models can tackle competition-level math problems. However, the success of such models hinges on access to massive high-quality math pre-training datasets—e.g., DeepSeekMath’s 120B tokens~\citep{shao2024deepseekmath} and Qwen-2.5-Math’s 1T tokens~\citep{yang2024qwen25math}. Yet no open-source dataset currently matches this scale and quality (see Table~\ref{app-tab:comparison_with_exisingt_corpora_statistics} for comparison), hindering progress on open math models.

A key obstacle lies in the limitations of current math web data pipelines. While web data forms the backbone of modern pre-training corpora~\citep{penedo2024fineweb, txt360data2024}, existing math-specific pipelines often suffer from overly aggressive pre-filtering (\emph{e.g.}, filtering based on HTML math tags \citep{pasteropenwebmath}), which causes many math-relevant documents to be missed. Moreover, widely used general-purpose text extraction tools are not optimized for mathematical content—they often strip or discard equations and symbols, severely degrading data quality~\citep{han24infimm,lozhkov2024finemath}. As a result, web-collected math data often lacks both scale and fidelity. Beyond web data, math-related code corpora (\emph{e.g.}, AlgebraicStack~\citep{azerbayev2023llemma}, MathCode-Pile~\citep{lu2024mathcoder2}) and synthetic datasets (\emph{e.g.}, WebInstruct~\citep{yue2024mammoth2}) have shown promising potential, but remain either limited in scale or not fully open-sourced.

\begin{figure}[!t]
  \centering
  \begin{minipage}{0.52\textwidth}
    \centering
    \vspace{-11pt}
    \includegraphics[width=0.95\linewidth]{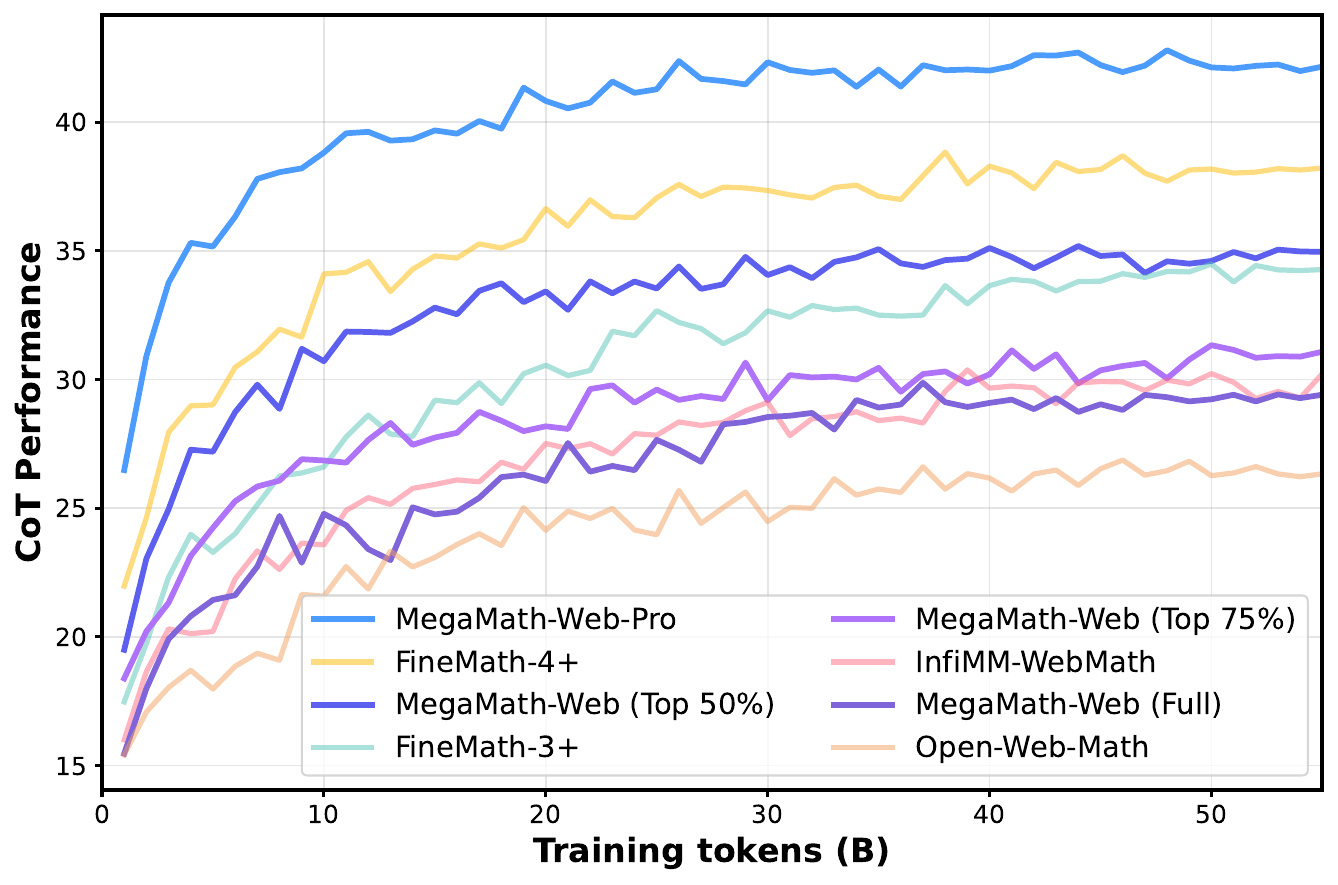}
    \vspace{-2mm}
    \end{minipage}
    \hspace{0.5mm}
\begin{minipage}{0.43\textwidth}
    \centering
    \vspace{-16pt}
    \includegraphics[width=0.95\linewidth]{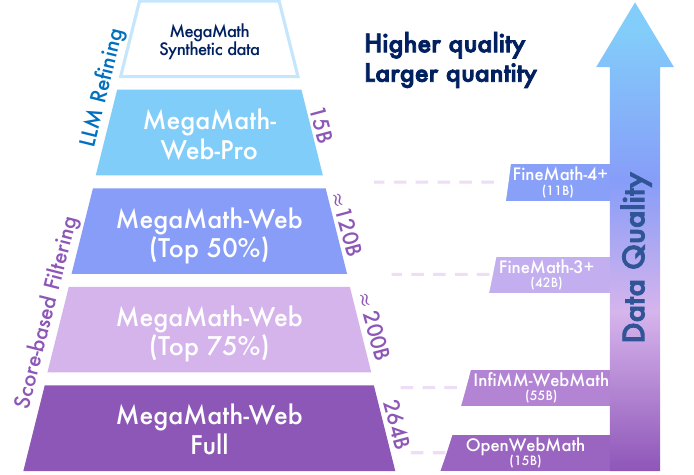}
    \vspace{-2mm}
    \end{minipage}
    \caption{%
    Comparison with existing open math corpora and \finemath-Web subsets.
    }\label{fig:megamath_web_comparison_with_existing_corpora}
    \vspace{-6.5mm}
\end{figure}

To bridge this gap, we introduce \finemath{} --- the largest open-source English math corpus to date, totaling 371.6B tokens. It comprises 279B tokens of web data, 28.1B of code, and 64.5B of synthetic data. During its construction, we conducted extensive ablation studies and optimizations across all domains to ensure both scalability and quality. For the web domain, we designed a two-stage, coarse-to-fine extraction and filtering pipeline, improving on the common pipeline. We reformatted math elements in HTML into compatible text representations (\emph{i.e.}, \LaTeX) to preserve equations and symbols during extraction. In the first stage, we applied a fast text extractor alongside a \fasttext{} classifier to filter candidate math documents. After deduplication, we reprocessed the retained HTMLs using a slower, high-quality extractor, followed by a second-stage \fasttext{} trained on the seed data from the first stage to mitigate distributional shift. This pipeline achieves both scale and fidelity, resulting in \finemath-Web. Based on this foundation, we further developed {\finemath-Web-Pro}, a premium subset delivering top quality via LM-based filtering and LLM refining, particularly beneficial for later training stages requiring higher data quality~\citep{hu2024minicpm}. In the code domain, we fine-tuned a small language model to filter math-relevant code snippets at scale, yielding \finemath-Code. For \finemath-Synthetic, we extracted and refined QA pairs from math web documents, translated non-Python code snippets into Python, and generated interleaved text-code samples from web content. Together, these efforts form a diverse and scalable math dataset backed by extensive empirical pre-training.

Our contribution can be summarized with following offerings in \finemath:
\vspace{-2.5mm}

\begin{enumerate}[leftmargin=1.25em,itemindent=0.15em,labelsep=0.2em,itemsep=0.05em]

\item An open math-focused dataset containing 371B tokens with optimized data curation pipelines, and a variety of data variants to cater to customized demands.
(\S\ref{subsec.curating_web} - \S\ref{subsec.final_data})
\item A comprehensive set of studies and ablation experiments that rigorously evaluate key design choices in the data accumulation process. (\S\ref{subsec.setup} - \S\ref{subsec.synthesis_ablation})
\item Empirical demonstrations including head-to-head comparison with existing math datasets~(\S\ref{subsec.existing_copora}, Figure~\ref{fig:megamath_web_comparison_with_existing_corpora}), and further training on latest \llamaiii series of models. (\S\ref{sec.put_together})
\end{enumerate}

\vspace{-3.5mm}

\section{\finemath Data Curation}
\label{sec.data_curation}

In this section, we will describe \finemath's whole data processing pipelines, which include three main components: web data~(\S~\ref{subsec.curating_web}), code data~(\S~\ref{subsec.curating_code}), and synthetic data generated from the former two~(\S~\ref{subsec.curating_synthesis}). 
Our key design choices are validated through downstream benchmarks or split validation sets. 
For computationally intensive operations like deduplication, we prioritize solutions that balance efficiency and effectiveness.

\subsection{Curating \finemath-Web}
\label{subsec.curating_web}
\begin{figure}[htbp]
\vspace{-2.5mm}
\centering
\includegraphics[width=0.85\linewidth]{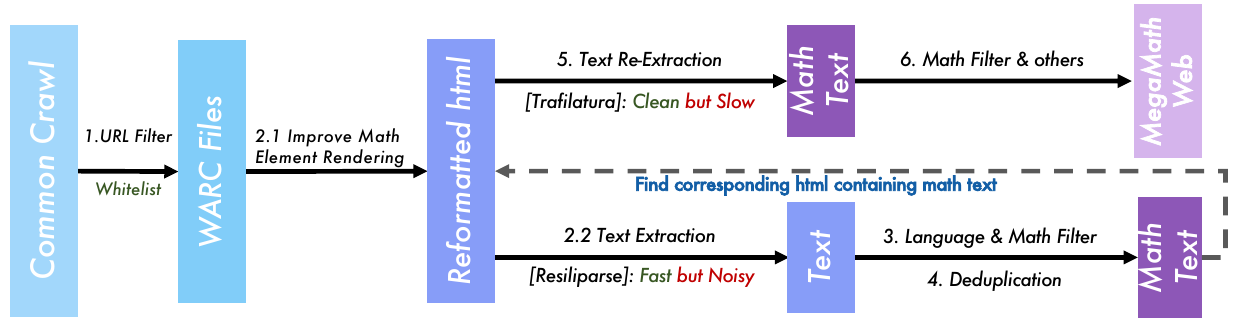}
\vspace{-2.5mm}
\caption{%
The pipeline for curating \finemath-Web from Common Crawl data.
}
\label{fig:cc_pipeline}
\vspace{-4.5mm}
\end{figure}

Web data takes up quite a lot of the general pre-training corpora, from which Common Crawl
is what has been widely used as pre-training data in many recent LLMs training~\citep{dubey2024llama3, yang2024qwen2, liu2024deepseek}.
In \finemath, we use 99 Common Crawl snapshots (\texttt{2014-15} to \texttt{2024-46}) as data source to extract high-quality math documents on the Internet.
The overall pipeline for web data is presented in Figure~\ref{fig:cc_pipeline}, with a detailed description in the following subsections.
In short, our pipeline contains the following steps: (1) \textit{data acquisition}; (2) \textit{first round text extraction}; (3) \fasttext{}\textit{-based math filtering}; (4) \textit{deduplication}; (5) \textit{second round text extraction}; (6) \textit{further filtering and post-processing}.

\subsubsection{Data Acquisition, URL Filtering, and Language Identification}
Instead of using \textbf{\textit{WET}}~(\textbf{W}ARC \textbf{E}ncapsulated \textbf{T}ext) data or simply filtering from public pre-training datasets~\citep{penedo2024fineweb, txt360data2024},
we re-extracted all the text from \textbf{\textit{WARC}}~(\textbf{W}eb \textbf{ARC}hive format) file format where each web page is stored as an HTML file.
This practice enables us to optimize text extraction from HTML, enhancing corpora quality specifically for math domain~(\S~\ref{subsubsec.text_extraction}).
We downloaded all available CC dumps and applied a URL filtering strategy before text extraction~\citep{penedo2023refinedweb} to exclude domains related to adult, gambling content, etc. 
Next, we used an off-the-shelf \fasttext{} model~\citep{joulin2016bag} for language identification and retained only English documents~(\texttt{score}$\geq0.65$).

\vspace{-3.5mm}
\subsubsection{Improved Text Extraction for Math Content}
\label{subsubsec.text_extraction}

Extracting texts from WARC using common extractors (e.g., \resili and \trafil) could produce higher-quality corpora over WET extraction~\citep{li2024datacomplm}. 
However, these extractors often fail to preserve math symbols and equations, even omitting them entirely~\citep{lozhkov2024finemath}. 
To address this, we introduced several HTML parsing optimizations specifically for \textbf{optimizing math expressions before extraction}. 
Our approach involved traversing the HTML DOM tree parsed by \resili and modifying math elements-related nodes to obtain an improved HTML file easy for text extraction including:
\begin{tcolorbox}[
colback=lightgray!10, 
colframe=black!60, 
boxsep=-1.pt
]
\begin{enumerate}[leftmargin=0.25em,itemindent=0.25em,labelsep=0.4em,itemsep=0.15em]
\item \textbf{Math Element Conversion:} Converts MathML and KaTeX content into LaTeX by extracting annotation tags or using mathml2latex~(with namespace handling) and recursively parsing HTML to accurately extract subscripts and superscripts.
\item \textbf{LaTeX Standardization and Transformation:} Removes unnecessary style commands, fixes symbol formatting issues, and converts HTML tags (such as \verb|<sup>|, \verb|<sub>|, and ``intbl'' spans) into appropriate LaTeX constructs.
\item \textbf{Unicode and Entity Conversion:} Maps mathematical Unicode characters and HTML entities to their corresponding LaTeX commands using W3C standards.
\end{enumerate}
\end{tcolorbox}
\paragraph{Two-stage Extraction} Our extraction process consisted of two phases, each serving a distance purpose. 
In practice, \resili and \trafil are widely used for pre-training corpora construction, but they have trade-offs: \resili is significantly faster and retains HTML elements more faithfully, while \trafil, though slower, removes noise more aggressively using various extraction engines and heuristics.
Unlike prior works based solely on \resili~\citep{pasteropenwebmath,han24infimm}, our pipeline first applied \resili for rapid extraction and filtering, significantly shrinking the candidate data size. 
For these candidate data, we then used \trafil on their WARC files for a second round HTML optimizations and text extraction, obtaining cleaner mathematical data. 
This coarse-to-fine approach improves text quality while maintaining development efficiency.

\subsubsection{Robust Math Document Recall}
Common Crawl~(CC) contains a vast array of texts from diverse domains. 
To effectively filter texts at scale, we require a robust and efficient classifier. 
We used \fasttext{}~\citep{bojanowski2017enriching}, a lightweight n-gram model, to score and identify math-related texts. 
During development, we identified the following key factors to obtain a robust \fasttext{} classifier:
\begin{tcolorbox}[
colback=lightgray!10, 
colframe=black!60, 
boxsep=-1.pt
]
\begin{enumerate}[leftmargin=0.25em,itemindent=0.25em,labelsep=0.4em,itemsep=0.15em]
\item \textbf{Text normalization}: Techniques like tokenization, case folding, digit normalization, and Unicode handling while managing whitespace and special characters achieve better training compatibility.
\item \textbf{Seed data}: Uniform sampling from Common Crawl and adding CoT data helps.
\item \textbf{Comprehensive evaluation}: Expanding beyond web texts to Wikipedia, textbooks, StackExchange and research papers improves recall assessment.
\end{enumerate}
\end{tcolorbox}
\paragraph{\fasttext{} Training} We started \fasttext{} training with one million positive and negative seed documents from \owm and random web documents from CC. Initially, we used a single snapshot dump for development, which risked reinforcing biases. 
To mitigate this, we sampled from all CC dumps and retrained the classifier during the second-round filtering process. We used Llama-3.1-70B-Instruct~\citep{dubey2024llama3} to automatically annotate math relevance scores (see Figure~\ref{prompt:web_math_score} for the prompt) on these filtered documents and CoT data was incorporated into the positive set as well, resulting in two million seed data. We used the same training hyperparameters as \dsmath{}~\citep{shao2024deepseekmath}.
\vspace{-2.5mm}
\paragraph{\fasttext{} Evaluation} 
When iterating training strategy, we found evaluation on 20K in-distribution~(ID) samples yielded easily \textbf{over 90\%} F1 score, masking \fasttext{}'s true performance. 
We thus created an out-of-distribution (OOD) suite by sampling arXiv, StackExchange, Wikipedia, and Textbook data from \mathpile~\citep{wang2024mathpile}.
In the OOD setting, our text normalization and training adjustments boosted the average F1 score from \textbf{81.8\%} to \textbf{98.8\%}, validating the effectiveness of our training strategy.
\vspace{-2.5mm}
\subsubsection{Data Deduplication}
Data deduplication plays a vital role in data curation process, especially for improving training efficiency, stability and reducing data memorization~\citep{lee2021deduplicating,tokpanov2024zyda}.

We adopted the Locality Sensitive Hash~(LSH) implementation of MinHash~\citep{broder2000identifying} for efficiency. 
Given two documents, the probability that they are assigned to the same hash bucket depends on their Jaccard similarity~\citep{broder1997resemblance} \(S\) and is given by
\(
    P = 1 - (1 - S^b)^r
\)
where \(b\) denotes the number of hash functions per bucket and \(r\) represents the number of buckets. 
Given a fixed hash permutation scheme (\(b \times r\)), which is strongly correlated with memory cost, and a target Jaccard similarity threshold \(t\), it is desirable to find the optimal deduplication configuration—one that ensures a rapid decay of \(P\) for any \(S \leq t\).
Considering our CPU capacity, we evaluated multiple configurations with 
the number of permutations between 110 and 128 and \(t \in \{0.70, 0.75, 0.80\}\). 
Assisted by training experiments, we determined that the most feasible choice is \(r=11\), \(b=10\), and \(t=0.75\). 
\vspace{-1.5mm}
\subsubsection{Curating  \finemath-Web-Pro: A Premium Subset}
It is increasingly common practice to filter top-quality data due to its superior impact on model performance~\citep{abdin2024phi}. 
High-quality data not only enhances performance but also does so at a lower cost, making it ideal for continual pre-training, mid-training, or scenarios with limited budgets. 
We thus further developed \finemath-Web-Pro, a premium subset filtered and refined from \finemath-Web. 
We employed the FineMath classifier~\citep{lozhkov2024finemath} to filter out low-quality text.
Subsequently, we used LLMs to further refine the text,\textbf{ ultimately delivering 15.1B tokens that significantly surpass all existing math corpora such as FineMath-4plus~\citep{lozhkov2024finemath}~(cf. Figure~\ref{fig:megamath_web_comparison_with_existing_corpora})}.
Though LLM was involved, we focused primarily on noise removal and text reorganizing thus this is not categorized as pure synthetic data.
See \S\ref{app-sec:pro-subset} for full developing strategy.

\vspace{-1.5mm}
\subsection{Curating \finemath-Code}
\label{subsec.curating_code}
\begin{figure}[htbp]
\centering
\vspace{-12.5pt}
\includegraphics[width=0.82\linewidth]{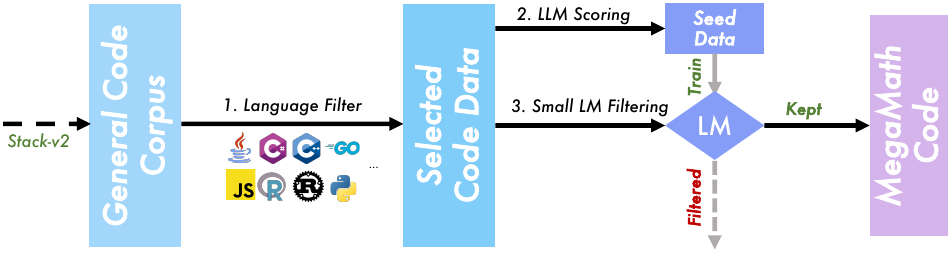}
\vspace{-1.5mm}
\caption{%
The pipeline for curating \finemath-Code.
}
\label{fig:stack_pipeline}
\end{figure}
Code pre-training has proved to enhance general reasoning~\citep{shao2024deepseekmath, aryabumi2024code}, and LLMs have also shown great potential to leverage code for problem-solving~\citep{goutora, li2024numinamath}. 
Thus, we believe blending code in math LLM training is also crucial.
We built \finemath{}-Code based on the Stack V2~\citep{lozhkov2024starcoder2}, and
employed a multi-step pipeline to recall high-quality code relevant to mathematical reasoning, logic puzzles, and scientific computation. 
As shown in Figure~\ref{fig:stack_pipeline}, our pipeline consists of:  
(1) \textit{Programming Language Selection} (\S\ref{subsubsec.pl_selection}) and  
(2) \textit{SLM-based Code Recall} (\S\ref{subsubsec.pl_filter}).  
\subsubsection{Programming Language Selection} \label{subsubsec.pl_selection}
The code pre-training corpus includes hundreds of programming languages; 
however, many of these languages are primarily associated with domains that are not closely related to mathematics or scientific computation~\citep{lozhkov2024starcoder2}.
In order to reduce the cost of model-based recall, we selected eleven programming languages based on choices made in previous studies~\citep{azerbayev2023llemma,xu2024lemur}: \texttt{C, C\#, C++, Go, Java, JavaScript, Python, R, Rust, Shell, SQL}. 
The selected languages are either extensively used in scientific computing and numerical operations or represent a significant portion of the corpus, which may potentially include mathematics-related snippets.

\subsubsection{SLM-based Code Data Recall}
\label{subsubsec.pl_filter}
We applied a small language model (SLM) based recall mechanism to identify math-related code snippets from public code pre-training datasets.
Inspired by recent works~\citep{penedo2024fineweb,zhou2024programming,wei2024arctic}, we first used a strong LLM to score code quality~(educational value) and mathematical relevance, assigning a discrete score from 0 to 5 for each aspect, as applied in several works~\citep{yuan2024self,penedo2024fineweb}.
Then, we trained a SLM on these data for large-scale filtering. (Please see \S\ref{app:code} for more details).
We also found that:
(1) \textbf{Stricter filtering greatly enhances performance to solve problems using code};
(2) \textbf{Allocating no more than 20\% of code data maximizes code-integrated problem-solving ability while maintaining NL reasoning benefits}. 
This aligns with DeepSeekMath's~\citep{shao2024deepseekmath} training recipe and further reinforces the justification for our filtering strategy, and we also empirically show the reasonability of this choice in \S\ref{subsec.code_ablation}.

\subsection{Curating \finemath-Synthetic Data}
\label{subsec.curating_synthesis}

Beyond being a high-quality mathematical corpus, \finemath also serves as \textbf{a strong foundation for large-scale data synthesis}. 
We explored data synthesis methods to further enhance both the quantity and quality of our dataset. 
Our synthesis spans three distinct formats: (1) \textit{Q\&A data}, (2) \textit{code data}, and (3) \textit{text \& code block data}.

\begin{figure}[htbp]
\centering
\includegraphics[width=0.85\linewidth]{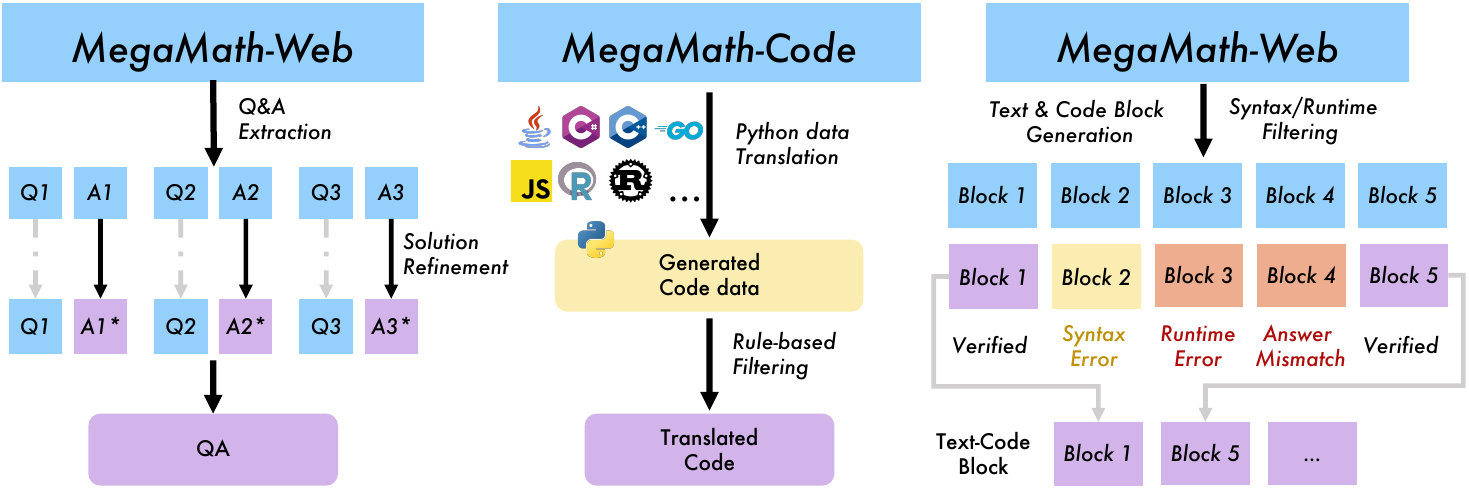}
\caption{%
The pipeline for curating synthetic data. \textbf{Left}: QA data generation; \textbf{Middle}: Python code augmentation; \textbf{Right}: text \& ode block data curation.
}
\label{fig:synthetic_pipeline}
\vspace{-4.5mm}
\end{figure}

\paragraph{Q\&A Extraction}
\label{subsubsec:text-data-synth}
Question-and-answer data is inherently well-structured and embodies a concentrated form of knowledge, making it valuable for problem-solving benchmarks~\citep{maini-etal-2024-rephrasing}.
Recent work reveal that these data can be found in pre-training data with massive quantity~\citep{yue2024mammoth2}.
We thus integrate and further verify this in \finemath. 
Our pipeline contains two steps: 
(1) identify and extract Q\&A pairs from the raw documents; 
(2) refine the Q\&A to make up or improve the intermediate reasoning steps.

To improve diversity and accumulate quantity, we ensembled refined Q\&A data from Qwen-2.5-72B-Instruct~\citep{yang2024qwen2} and Llama-3.3-70B-Instruct~\citep{dubey2024llama3}.

\paragraph{Code Translation}
To enhance Python code data, we employed LLMs to ``translate'' code from other programming languages into Python, thereby augmenting the code data. 
We adopted a straightforward zero-shot prompting approach using open-source LLMs or code-specialized LLMs to enhance the volume of Python code. Specifically, we experimented with two models: Qwen2.5-Coder-32B-Instruct~\citep{hui2024qwen2} and Llama-3.1-70B-Instruct~\citep{dubey2024llama3}.
Please kindly refer to \S\ref{app:code} for the translation prompt.
\paragraph{Text \& Code Block Generation}
Recent work by \citet{lu2024mathcoder2} introduced a synthesis pipeline for obtaining \textit{``generated mathematical code''}, consisting of interleaved text, symbolic expressions and code blocks. 
Such data has been shown to enhance a model’s ability to generate Python snippets for solving mathematical problems and to leverage execution feedback to refine solution steps.  
As illustrated in Figure~\ref{fig:synthetic_pipeline}, we unify such process into:  
(1) \text{LLM-based generation}: given a document, LLMs generate multiple structured blocks including title, mathematical expression, result, and corresponding code;  
(2) \text{Verification via execution}: ensures that the generated code executes correctly without errors and produces expected outputs; 
(3) \text{Packing verified blocks}: combines validated blocks into a single training sample for downstream use.  
Besides, we found \citet{lu2024mathcoder2}
did not account for malicious code, handling only basic errors like timeouts.
To address these issues, we implemented a pre-filtering mechanism based on Abstract Syntax Tree (AST).
Any snippet flagged as risky is excluded from execution, safeguarding a 100\% execution success rate without abrupt halts or segmentation faults during our curation.
\vspace{-2.5mm}
\subsection{Dataset Decontamination}
\label{subsec.decont}
To mitigate benchmark contamination~\citep{xu2024benbench}, we checked the overlap between \finemath and 12 downstream benchmarks widely used in evaluating LLM's mathematical reasoning ability such as GSM8K~\citep{cobbe2021training-gsm8k}, MATH~\citep{hendrycks2measuring-math}, MMLU~\citep{hendrycks2020measuring-mmlu}, and AIME~\footnote{\href{https://huggingface.co/datasets/AI-MO/aimo-validation-aime}{https://huggingface.co/datasets/AI-MO/aimo-validation-aime}}.
We concatenated problems and solutions together as a whole sample, checked the exact 13-gram match, and ruled out contaminated documents. 
This further removes about \textbf{0.01\%} of the documents from the dataset.
\vspace{-2.5mm}
\subsection{The Final Dataset: \finemath 371B Collection}
\label{subsec.final_data}
\vspace{-2.5mm}
\begin{table}[!t]
    \centering
    \setlength{\tabcolsep}{12.5pt}
    \definecolor{megablue}{HTML}{CBEFFF}
    \caption{The category and statistics of \finemath.}
    \small
    \label{tab:megamath_statistics}
    \resizebox{0.75\textwidth}{!}{%
        \begin{tabular}{l r r r}
            \toprule
            \textbf{Category} & \textbf{\# Sample(M)} & \textbf{\# Toks(B)} & \textbf{Avg. (\# Toks)} \\
            \midrule 
            \rowcolor{megablue}
            \textbf{Web Domain}   &  \textbf{121.5} & \textbf{279.0} & \textbf{2296.9} \\
            \quad Web & 106.5  & 263.9  & 2478.7 \\
            \quad Web-Pro & 15.0  & 15.1  & 1006.0  \\
            \rowcolor{megablue}
            \textbf{Code Domain} & \textbf{13.4} & \textbf{28.1}  & \textbf{2102.7} \\
            \midrule
            \rowcolor{megablue}
            \textbf{Synthetic Data} & \textbf{80.2} & \textbf{64.5} & \textbf{804.5} \\
            \quad \text{Translated Code}        & 7.4  & 7.2  & 979.5 \\
            \quad \text{Q\&A}        & 22.6  & 7.0  & 308.3 \\
            \quad \text{Text\&Code Block}  & 50.2  & 50.3 & 1002.1 \\
            \midrule
            \rowcolor{megablue}
            \textbf{Total} & \textbf{215.1} & \textbf{371.6}  & \textbf{1727.6} \\
            \bottomrule
        \end{tabular}
    }
\end{table}

Combining all previous efforts together, the final collection of \finemath datasets currently contained a total of {$371$B} tokens~(count by the Llama-2 tokenizer).
We present a detailed breakdown statistics about \finemath in 
\textbf{Table~\ref{tab:megamath_statistics}}.
Designed for various training stages, training budgets, and base model capability,
we offer a collection of \finemath data variants including:
(1) \textcolor[HTML]{0070C0}{\textbf{\finemath-Web}}: the complete web dataset consisting of 263.9B tokens, and also \textcolor[HTML]{0070C0}{\textbf{\finemath-Web-Pro}} (15.1B), the top-quality subset obtained through LM-based scoring and refining.
(2) \textcolor[HTML]{0070C0}{\textbf{\finemath-Code}} (28.1B): math-related code corpus recalled from Stack-v2.
(3) \textcolor[HTML]{0070C0}{\textbf{\finemath-Synth}} (64.5B): LLM-based synthetic data enhancing both the quality and quantity, covering three distinct formats of text and code data.
\vspace{-2.5mm}

\section{Ablation and Demonstration of \finemath at Scale with Pre-training}
\label{sec.ablation}

During data curation, we conducted extensive pre-training experiments on \finemath to ablate each key decision. 
In this section, we present the experimental details, key results, and finally scale up training to further demonstrate the effectiveness of \finemath. 

\subsection{Setup}
\label{subsec.setup}
\paragraph{Proxy LM for Ablation}
During development, we used a small proxy model for ablations on each data source and component. 
We chose TinyLlama-1B~\citep{zhang2024tinyllama} for its small size and transparent training, ensuring it effectively monitors data quality. 
We trained within a controlled budget, typically set to \textbf{5/15/55} B tokens, depending on dataset size and experimental cost, and evaluated performance at 1B token intervals.
\vspace{-2.5mm}
\paragraph{Evaluation}
We used a total of 10 math-related benchmarks, splited into two sets: \textbf{\texttt{Core}} and \textbf{\texttt{Extended}}. The \textbf{\texttt{Core}} set includes five math-focused tasks with stable improvements even under limited training, such as GSM8K and MATH. Building on this, the \textbf{\texttt{Extend}} set further includes five datasets, either indirectly related to math or with performance fluctuations, such as MMLU-STEM. We employ two prompting-based evaluations: (1) few-shot CoT~\citep{wei2022chain} for all benchmarks; (2) PAL~\citep{gao2023pal} for the \textbf{\texttt{Core}} set to assess problem-solving via Python code generation.
Please check \S~\ref{app-sec:eval} for more details.
\subsection{Ablation on \finemath-Web}
\vspace{-0.2cm}
\begin{wraptable}[8]{r}{0.49\textwidth}
\vspace{-4.5mm}
\centering
\caption{Ablation on Text Extraction for Math
}
\vspace{-2.5mm}
\small
\definecolor{megablue}{HTML}{CBEFFF}
\renewcommand{\arraystretch}{0.8}
\scalebox{0.95}{
\begin{tabular}{>{\centering\arraybackslash}m{1.8cm} | >{\centering\arraybackslash}m{1.8cm}| >{\centering\arraybackslash}m{0.8cm} >{\centering\arraybackslash}m{0.8cm}}
\toprule
\textbf{Text Extractors} & \textbf{\begin{tabular}[c]{@{}c@{}} w/ HTML \\ Optimization\end{tabular}} & \textbf{Core Avg.}                & \textbf{Ext. Avg.}            \\ \midrule
\textit{\color[HTML]{9B9B9B} Base Model}                  & -                 & {\color[HTML]{9B9B9B} 11.2} & {\color[HTML]{9B9B9B} 14.7} \\ \midrule
\trafil        & \ding{55}              & 22.0                        & 19.2                          \\
\resili         & \ding{52}              & 22.5                       & 18.6                        \\
\rowcolor{megablue}
\trafil         & \ding{52}              & \textbf{23.8}              & \textbf{20.6}                    \\ \bottomrule
\end{tabular}
\label{tab:text_extraction_ablation}
}
\end{wraptable}
\paragraph{Importance of optimizing text extraction for math content} We conducted continual pre-training experiments within a 15B-token training budget on one dump from 2024. The training corpora consisted of filtered math documents from vanilla \trafil, and text extracted from the optimized HTML using \resili and \trafil, all derived from the first-round filtering. 
During original \trafil's extraction, \texttt{<math>} elements in HTML were directly discarded. 
After applying specialized optimizations for math-related HTML, the extracted data from \trafil well-preserved math symbols and clearly improved CoT downstream performance~(cf.~Table~\ref{tab:text_extraction_ablation}). When both extractors operated on our optimized HTML, \resili preserved more noise from the original documents, leading to lower data quality compared to \trafil.
\begin{wraptable}[9]{r}{0.460\textwidth}
\centering
\caption{Ablation on MinhashLSH Dedup.}
\vspace{-2.5mm}
\small
\definecolor{megablue}{HTML}{CBEFFF}
\scalebox{0.9}{
\begin{tabular}{>{\centering\arraybackslash}m{1.1cm} | >{\centering\arraybackslash}m{0.4cm} >{\centering\arraybackslash}m{1.4cm} | >{\centering\arraybackslash}m{0.7cm} >{\centering\arraybackslash}m{0.7cm}}
\toprule
\textbf{(r, b)} & \textbf{t} & \textbf{Tokens Left (B)} & \textbf{Core Avg.} & \textbf{Ext. Avg.} \\ \midrule

(14, 9)         & 0.70       & 16.0                     & 17.3               & 16.6               \\
(14, 8)         & 0.75       & 23.5                     & 19.1               & 17.0               \\
\rowcolor{megablue}
\textbf{(11, 10)} & \textbf{0.75} & \textbf{26.0}        & \textbf{19.4}      & \textbf{17.5}      \\
(11, 11)        & 0.75       & 25.0                     & 19.2               & 16.0               \\
(9, 12)         & 0.80       & 29.0                     & 18.8               & 16.9               \\
(9, 13)         & 0.80       & 30.0                     & 17.6               & 15.7               \\ \bottomrule
\end{tabular}
}
\label{tab:dedup_ablation}
\end{wraptable}
\paragraph{Parameters of Deduplication} To minimize redundancy and reduce the costs associated with follow-up text re-extraction using \trafil, we conducted ablation pre-training experiments on all 2014 dumps with a 55B-token training budget to optimize the parameters for Minhash LSH. Our goal was to preserve downstream CoT performance while retaining as many mathematical documents as possible within our cluster capacity. As shown in Table~\ref{tab:dedup_ablation}, applying \(r=11\), \(b=10\) provided the optimal balance. We reported the average of the last 5 checkpoints to avoid result fluctuations.
\vspace{-5.5mm}

\begin{wrapfigure}[13]{r}{0.40\textwidth}
\vspace{-4.5mm}
    \centering
    \begin{tikzpicture}
    \tiny
        \begin{axis}[
            xlabel={\textbf{Training Tokens (B)}},
            ylabel={\textbf{CoT Core Performance}},
            xmin=1, xmax=5,
            ymin=16, ymax=30,
            xtick={1,2,3,4,5},
            ytick={16, 22, 28},
            tick label style={font=\tiny},
            legend style={font=\tiny, draw=none, fill=none, at={(0.5,-0.15)}, anchor=north, legend columns = 2, align=right},
            grid=major,
            width=5.8cm,
            height=4.9cm
        ]
        \definecolor{colorwarmorange}{HTML}{9554BD} 
        \definecolor{colorwarmgreen}{HTML}{DBB2F0}   
        \definecolor{colorwarmblue}{HTML}{819EFF} 
        \definecolor{colorwarmred}{HTML}{66CFFE}   

        \addplot[color=colorwarmorange, mark=square, very thick] coordinates {
            (1, 16.62) (2, 19.35) (3, 20.80) (4, 21.81) (5, 22.45)
        };
        \addlegendentry{V1: OWM}
        
        \addplot[color=colorwarmgreen, mark=square, very thick] coordinates {
            (1,22.33) (2,25.74) (3,26.00) (4,27.29) (5,27.15)
        };
        \addlegendentry{V2: Random}

        \addplot[color=colorwarmblue, mark=triangle, very thick] coordinates {
            (1,21.76) (2,23.60) (3,26.46) (4,26.90) (5,27.82)
        };
        \addlegendentry{V2: Balance}

        \addplot[color=colorwarmred, mark=o, very thick] coordinates {
            (1,23.40) (2,25.77) (3,27.67) (4,28.59) (5,28.89)
        };
        \addlegendentry{V2: Balance + CoT data}

        \end{axis}
    \end{tikzpicture}
    \vspace{-4mm}
    \caption{Ablation on \fasttext{}}
    \label{fig:ablation_on_different_fasttext}
\end{wrapfigure} 
\paragraph{Ablation on \fasttext{}} Initially, we employed \owm{} as the positive seed data for training \fasttext{} used for the first round filtering with a loose threshold. While it worked well on the single dump used for initial development, it became less accurate when scaled to all dumps, likely due to shifting data distributions. We thus re-trained \fasttext{} with LLM-annotated math-related documents from all dumps as the positive seed data. 
As shown in Figure~\ref{fig:ablation_on_different_fasttext}, the re-trained \fasttext{} (V2) performed better in the second-round filtering compared to the initial version (V1). 
Our controlled experiments revealed that balanced sampling seed data from each dump provides a slight improvement while incorporating CoT data into positive seed data yielded significant gains. 
Note that the ablations were conducted on the top 10\% scoring filtered data from all dumps in 2024 within a 5B-token training budget. 
We further validated our decision through experiments on all dumps yearly as shown in Figure~\ref{app-fig:ablation_on_different_fasttext_all_year_appendix}.

\subsection{Ablation on \finemath-Code}
\label{subsec.code_ablation}
We conducted two sets of ablation studies using 5B training tokens: (1) evaluating the impact of recall filtering criteria on downstream performance~\footnote{When testing filtering criteria, we prioritize PAL results, and use Python subset for training.}, and (2) ensuring the recalled code dataset is sufficiently large for training despite aggressive filtering.
As shown in Table~\ref{tab:cot_pal_avg_filter_comparison}, stricter filtering (i.e., $S_\text{edu}\geq4$ and $S_\text{math}\geq4$) selects code data that significantly boosts PAL performance. 
Moreover, even though such strict filtering excludes many samples, the remaining data appears sufficient for effective mathematical training; in fact, mixing too much code data seems harmful for CoT Performance~(\emph{e.g.}, $\geq 30\%$).
\vspace{-2.5mm}
\begin{table}[htbp]
  \centering
  \begin{minipage}{0.45\textwidth}
    \centering
    \definecolor{megablue}{HTML}{CBEFFF}
    \caption{Ablation of filtering criteria.}
    \vspace{-2.0mm}
    \resizebox{\textwidth}{!}{%
    \small
      \begin{tabular}{c|cc}
      \toprule
      \textbf{Filter Criteria} & \textbf{CoT Avg.} & \textbf{PAL Avg.} \\
      \midrule
      \textbf{text only} & 19.0  & 15.6 \\
      {$S_\text{edu}\geq3, S_\text{math} \geq 3$} & 19.4  & 16.1 \\
      {$S_\text{edu}\geq3, S_\text{math} \geq 4$} & 19.8  & 16.8 \\
      {$S_\text{edu}\geq4, S_\text{math} \geq 3$} & 19.7  & 17.5 \\
      \rowcolor{megablue}
      {$S_\text{edu}\geq4, S_\text{math} \geq 4$} & 18.8  & 19.5 \\
      \bottomrule
      \end{tabular}%
    }
    \label{tab:cot_pal_avg_filter_comparison}
  \end{minipage}
  \hspace{0.75cm}
  \begin{minipage}{0.43\textwidth}
    \centering
    \caption{Ablation of data mixture ratios.}
    \vspace{-2.0mm}
    \definecolor{megablue}{HTML}{CBEFFF}
    \resizebox{\textwidth}{!}{%
    \small
      \begin{tabular}{c|cc}
      \toprule
      \textbf{Mix Ratio} & \textbf{CoT Avg.} & \textbf{PAL Avg.} \\
      \midrule
      \textbf{text only}   & 19.0  & 15.6 \\
      \textbf{code : text} = 12.5\%  & 19.3  & 17.4 \\
      \rowcolor{megablue}
      \textbf{code : text} = 20.0\%  & 19.5  & 18.4 \\
      \textbf{code : text} = 33.3\%  & 16.4  & 17.5 \\
      \textbf{code : text} = 50.0\%  & 17.5  & 18.8 \\
      \bottomrule
      \end{tabular}%
    }
    \label{tab:cot_pal_avg_mix_comparison}
  \end{minipage}%
\vspace{-3.5mm}
\end{table}

\subsection{Ablation on \finemath-Synthesis}
\label{subsec.synthesis_ablation}
Our synthesis development proceeds in parallel with web data acquisition. Thus, in web data synthesis, we started with existing public corpora rather than \finemath-Web. In particular, we utilized quality-filtered subsets of \owm{} and \infimm{} to lower experimental cost.
We focus on:
(1) Verify that the generated data can boost performance more effectively.
(2) Evaluate the impact of different prompts and models on performance, ultimately guiding better strategies.
\vspace{-1.5mm}

\begin{wraptable}[10]{r}{0.5\textwidth}
\centering
  \small
  \definecolor{megablue}{HTML}{CBEFFF}
  \caption{Ablations on prompt and comparison with other data. \textbf{FM-4plus}: FineMath-4plus.}
  \vspace{-2.5mm}
    \resizebox{0.45\textwidth}{!}{%
    \begin{tabular}{l|cc}
    \toprule
    \textbf{Data} & \textbf{Core Avg.} & \textbf{Ext. Avg.} \\
    \midrule
    \textbf{FM-4plus} & 28.3  & 19.6 \\
    \textbf{WebInstruct} & 34.6  & 17.6 \\
    \midrule
    \textbf{Vanilla Prompt} & 39.2  & 19.5 \\
    \quad \textit{w.} \text{ELI5} & 41.3  & 19.2 \\
    \rowcolor{megablue}
    \quad \textit{w.} \text{ELI5 + IC} & \textbf{48.8}  & \textbf{23.6} \\
    \bottomrule
    \end{tabular}%
}
  \label{tab:qa_ablation}%
\end{wraptable}
\paragraph{QA generation} We implemented the two-stage pipeline in WebInstruct~\citep{yue2024mammoth2} but using the latest and most capable LLMs. 
Through several prompting iterations, we found: (1) using an ELI5-style~(Explain like I am five) prompt for QA refining produces structured solutions; (2) emphasizing information completeness enhances data quality further.
Table~\ref{tab:qa_ablation} shows the 5B training results on different datasets: using prompt with ELI5 improves performance, and further adding information completeness (ELI5 + IC) yields the best results, with \texttt{Core} and \texttt{Extended} scores of \textbf{48.8} and \textbf{23.6}. 
These results indicate that structured and comprehensive extraction are keys to enhancing QA data, and also show that QA style data exhibits superior performance to web documents.

\begin{wraptable}[10]{r}{0.49\textwidth}
\vspace{-4.5mm}
      \centering
      \small
      \caption{Ablations on Code Synthesis: The default code-to-text ratio is 1:7; ``full'': no text is mixed. We exclude~\citet{lu2024mathcoder2} due to its partial release ($\approx$0.25B tokens).
      }
      \vspace{-2.5mm}
      \resizebox{0.48\textwidth}{!}{%
        \begin{tabular}{l|cc}
        \toprule
        \textbf{Data} & \textbf{CoT Avg.} & \textbf{PAL Avg.} \\
        \midrule
        \textbf{code} & 18.8  & 19.5 \\
        \textbf{trans. code} & 19.0 & 20.6 \\
        \textbf{text \& code block} & 22.5 & 28.1 \\
        \textbf{text \& code block} (full) & \textbf{30.8} & \textbf{46.5} \\
        \bottomrule
        \end{tabular}%
        }
      \label{tab:code_synth_ablation}%
\end{wraptable}
\paragraph{Code Synthesis}
Our experiments compared training on raw code data to methods incorporating code translation and interleaved text \& code blocks. With a controlled training budget of 5B tokens (see Table~\ref{tab:code_synth_ablation}), translated code (trans. code) yields modest downstream improvements over raw code, while adding code block data further enhances both CoT and PAL performance, even without mixing text data~(see the ``full'' line results).
Also, These results clearly show that synthetic data achieves higher quality.

\subsection{Comparison with Existing Math Corpora}\label{subsec.existing_copora}

To assess the data quality of \finemath, we performed continual pre-training on existing corpora within a 55B token budget. We compared \finemath-Web with mainstream large-scale corpora, including \owm{}, \infimm{}, and the latest FineMath release. As previously shown in \textbf{Figure~\ref{fig:megamath_web_comparison_with_existing_corpora}}, \finemath-Web already achieves corpus quality comparable to \infimm{} in downstream tasks but providing substantial more tokens, with performance improving if we use higher-scored subsets (top 75\% and top 50\%). Notably, \finemath-Web-Pro outperforms both FineMath-3+ and FineMath-4+ by {$\mathbf{\geq 4\%}$}, delivering the highest-quality corpus to date. Furthermore, these \finemath-Web variants show the potential to offer flexible options to accommodate different computing budgets.

\subsection{Putting It All Together: Training \finemath on Cutting-Edge LMs}
\label{sec.put_together}
\begin{wrapfigure}[13]{r}{0.45\textwidth}
\vspace{-4.5mm}
    \centering
    \includegraphics[width=\linewidth]{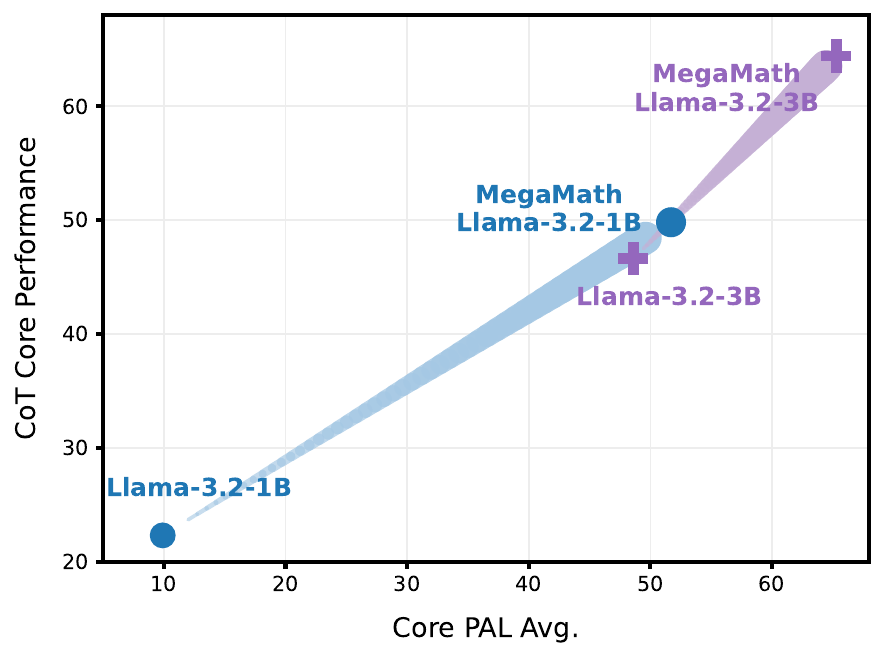}
    \vspace{-8.0mm}
    \caption{Training on \llamaiii.2-1B/3B.}
    \label{fig:megamath_put_together}
\vspace{-4.5mm}
\end{wrapfigure}
We demonstrate the effectiveness of \finemath{} by training it on state-of-the-art open LLMs—the \llamaiii.2 series. 
Given that the \llamaiii models have been extensively trained on 14.8T tokens~\citep{dubey2024llama3} and exhibit strong performance across various benchmarks, we believe they exemplify state-of-the-art capability and robust performance, making them an ideal validation point. 
For training, we adopt and refine the data mixture configurations from DeepSeekMath and Llemma to accommodate our diverse data sources, and train 100B tokens for \llamaiii.2-1B and 50B tokens for \llamaiii.2-3B.
We evaluate all models under CoT and PAL configurations. 
As shown in Figure~\ref{fig:megamath_put_together}, the MegaMath series of models achieves a 15\% to 20\% CoT performance improvement over Llama — for example, reaching 56.2\% on GSM8K and 25.1\% on MATH for the 3B model — with a similar boost observed on PAL. 
This clearly demonstrates the exceptional quality and effectiveness of \finemath in advancing mathematical reasoning in state-of-the-art language models.
Please refer to \S~\ref{app-sec:llama-train} and \S~\ref{app.subsec:full_ablations} for training configuration and full evaluation results.

\section{Related Works}
\vspace{-1.5mm}
\paragraph{Mathematical Pre-training Corpus and Syntheic Datasets}
OpenWebMath~\citep{pasteropenwebmath} curated its data from web pages, with strict filtering which may remove potential documents. 
MathPile~\citep{wang2024mathpile} diversified from web domains and built datasets mostly from arXiv papers and textbooks. 
Furthermore, InfiMM-Web-Math~\citep{han24infimm} assembled a multimodal dataset pairing math text with images.  
Recently, FineMath~\citep{lozhkov2024finemath} was developed by retrieving from FineWeb~\citep{penedo2024fineweb} and using a BERT classifier to select clear, step-by-step math explanations. 
For synthetic math datasets,
recent work such as NuminaMath~\citep{li2024numinamath} converted competition-level problems into chain-of-thought solutions via tool-assisted reasoning. Meanwhile, Skywork-Math~\citep{zeng2024skyworkmath}, OpenMathInstruct-2~\citep{toshniwal2024openmathinstruct} and WebInstruct~\citep{yue2024mammoth2} generated large-scale QA pairs from open benchmarks and web contents.
MathCoder2~\citep{lu2024mathcoder2} used a 19.2B-token MathCode-Pile combining filtered datasets with synthetic code data.
In \finemath, we aim to build a large-scale dataset that matches proprietary corpora via reproducible pipelines, diverse data sources, and thorough sanity checks, finally covering larger quantity and higher quality dataset.

\section{Conclusion and Future Work}
\vspace{-1.5mm}
We introduce \finemath, the largest training corpus to date tailored for the mathematical domain, comprising 371B tokens from web sources, code corpora, and synthesized data.
Comprehensive ablation studies guide us to efficient curation of high-quality, domain-specific datasets. 
Large-scale continual pretraining on \llamaiii series of model further demonstrates \finemath's effectiveness by producing strong math base models.
We hope the \finemath dataset, alongside our released artifacts, can foster further research in mathematical reasoning and domain-specific language modeling.

\clearpage
\newpage
\bibliography{colm2025_conference}
\bibliographystyle{colm2025_conference}

\newpage
\appendix

\section{Comparison with Existing Corpora}
\begin{table}[htbp]
\centering
\caption{Comparison with existing large-scale math corpora}
\resizebox{1.\textwidth}{!}{
\begin{tabular}{lcSccc}\toprule
\textbf{Corpus Name} &\textbf{Fully Public} &\textbf{\# Tokens (B)} &\textbf{Date} &\textbf{Type} \\\midrule
OpenWebMath & \ding{52}  &14.5 &2023 Oct. &Web \\
AlgebraicStack &\ding{52}  &11.0 &2023 Oct. &Code \\
MathPile & \ding{52}  &9.5 &2023 Dec. & \makecell{ArXiv, Web, Textbooks,\\ StackExchange, Wiki} \\
DeepseekMath & \ding{55}  &120 &2024 Feb. &Web \\
InfiMMWebMath & \ding{52}  &55.0 &2024 Sep. &Web \\
Qwen Math Corpus v2 & \ding{55} & 1000 & 2024 Sep. & \makecell{Web, Code snippets, Encyclopedias, \\ Books, Exam questions, Synthetic data} \\
MathCode-Pile & \ding{55} &19.1 &2024 Oct. &Web, Code, Textbooks \\
FineMath & \ding{52} &34 &2024 Dec. &Web \\
\midrule
\multicolumn{5}{c}{\textbf{\finemath Collection~(Ours)}} \\ \midrule
MegaMath-Web &\ding{52}  &263.9 &\multirow{6}{*}{2025 Apr.} &Web \\
MegaMath-Web-Pro & \ding{52}  & 15.1 & &Web \\
MegaMath-Code & \ding{52}  & 28.1 & &Code \\
MegaMath-Synth-Code & \ding{52}  & 7.2 & &Code \\
MegaMath-Synth-Q\&A & \ding{52}  & 7.0 & &Q\&A \\
MegaMath-Synth-Text\&Code & \ding{52}  & 50.3 & &Interleave text\&code \\
\bottomrule
\end{tabular}}
\label{app-tab:comparison_with_exisingt_corpora_statistics}
\end{table}

\section{Details for Curating \finemath-Web}

The scoring prompt for evaluating web documents' relevance to mathematics is presented in Figure~\ref{prompt:web_math_score}.
\begin{figure}[ht]
\small
\centering
\begin{tcolorbox}
\begin{Verbatim}[breaklines=true,breaksymbol=,]
Please evaluate the given document for its relevance to mathematics and assign a score from 0 to 5. Use the following scoring criteria:
5: The document is entirely about mathematics, containing numerous mathematical concepts, formulas, proofs, or advanced mathematical educational content.
4: The document is primarily about mathematics but may include some applications in other disciplines or content related to mathematics education.
3: The document contains significant mathematical content, but it's not the main focus. It might be mathematical applications in physics, engineering, or similar fields.
2: The document includes some mathematical elements, such as basic calculations, simple statistics, or graphs, but these are not the main content of the document.
1: The document has very little mathematics-related content, possibly only mentioning numbers or simple calculations in passing.
0: The document has no mathematical content whatsoever.

The document is given as:
<EXAMPLE>.

After examining the document: 
- Briefly justify your total score, up to 100 words.
- Conclude with the score using the format: "Score:  <total points>"
\end{Verbatim}
\end{tcolorbox}
\caption{Scoring Prompts for evaluating web documents relavance to mathematics.}
\label{prompt:web_math_score}
\end{figure}

\subsection{Fine-grained Deduplication}
\label{app-sec:fine-grained-dedup}
We also explored several fine-grained deduplication methods, including exact substring~\citep{lee2021deduplicating} and sentence-level deduplication~\citep{raffel2020exploring}. 
Initially, we found that removing duplicates disrupted text consistency. 
To mitigate this, we attempted trimming only the head and tail portions, but still identified many math expressions and degraded downstream performance. 
We suspect the effectiveness of these methods depends on text extraction techniques and may be more suitable for \resili. We thus leave this for future exploration.

\subsection{Strategy for \finemath-Pro Subset}
\label{app-sec:pro-subset}
\begin{wraptable}[12]{r}{0.55\textwidth}
\vspace{-3.5mm}
  \centering
  \caption{Yearly Ablation of Edu scoring strategy.}
\vspace{-2.0mm}
\resizebox{0.55\textwidth}{!}{%
    \begin{tabular}{c|cccccc}
    \toprule
    \multicolumn{1}{c}{} & \textbf{GSM8K} & \textbf{MATH} & \textbf{ASDiv} & \textbf{SVAMP} & \textbf{MAWPS} & \textbf{AVG} \\
    \midrule
    \textbf{FM-4plus} & 10.5  & 6.1   & 41.9  & 25.3  & 57.9  & 28.3 \\
    \midrule
    \rowcolor[rgb]{ 1,  .937,  .608} \textbf{2014} & \cellcolor[rgb]{ .878,  .902,  .588}6.0 & \cellcolor[rgb]{ .863,  .894,  .584}3.7 & \cellcolor[rgb]{ .796,  .875,  .569}30.2 & \cellcolor[rgb]{ .867,  .898,  .584}17.5 & \cellcolor[rgb]{ .871,  .898,  .584}35.4 & \cellcolor[rgb]{ .855,  .894,  .584}18.6 \\
    \rowcolor[rgb]{ 1,  .937,  .608} \textbf{2015} & \cellcolor[rgb]{ .937,  .918,  .6}5.0 & \cellcolor[rgb]{ 1,  .937,  .612}3.0 & \cellcolor[rgb]{ 1,  .937,  .612}21.8 & \cellcolor[rgb]{ 1,  .937,  .612}14.4 & \cellcolor[rgb]{ 1,  .937,  .612}27.7 & \cellcolor[rgb]{ 1,  .937,  .612}14.4 \\
    \rowcolor[rgb]{ 1,  .937,  .608} \textbf{2016} & \cellcolor[rgb]{ 1,  .937,  .612}3.9 & \cellcolor[rgb]{ .725,  .851,  .557}4.4 & \cellcolor[rgb]{ .839,  .886,  .58}28.4 & \cellcolor[rgb]{ .894,  .906,  .592}16.9 & \cellcolor[rgb]{ .867,  .898,  .584}35.8 & \cellcolor[rgb]{ .878,  .902,  .588}17.9 \\
    \rowcolor[rgb]{ .937,  .918,  .608} \textbf{2017} & \cellcolor[rgb]{ .855,  .894,  .584}6.4 & \cellcolor[rgb]{ .608,  .816,  .529}5.0 & \cellcolor[rgb]{ .678,  .839,  .545}34.9 & \cellcolor[rgb]{ .682,  .839,  .545}21.8 & \cellcolor[rgb]{ .718,  .851,  .553}44.6 & \cellcolor[rgb]{ .718,  .851,  .553}22.5 \\
    \rowcolor[rgb]{ .937,  .918,  .608} \textbf{2018} & \cellcolor[rgb]{ .867,  .898,  .584}6.2 & \cellcolor[rgb]{ .431,  .761,  .494}5.9 & \cellcolor[rgb]{ .686,  .839,  .545}34.6 & \cellcolor[rgb]{ .647,  .827,  .537}22.6 & \cellcolor[rgb]{ .682,  .839,  .545}46.7 & \cellcolor[rgb]{ .694,  .843,  .549}23.2 \\
    \rowcolor[rgb]{ .937,  .918,  .608} \textbf{2019} & \cellcolor[rgb]{ .855,  .894,  .584}6.4 & \cellcolor[rgb]{ .647,  .827,  .537}4.8 & \cellcolor[rgb]{ .608,  .816,  .529}37.7 & \cellcolor[rgb]{ .686,  .839,  .549}21.7 & \cellcolor[rgb]{ .651,  .827,  .541}48.4 & \cellcolor[rgb]{ .675,  .835,  .545}23.8 \\
    \rowcolor[rgb]{ .937,  .918,  .608} \textbf{2020} & \cellcolor[rgb]{ .722,  .851,  .553}8.7 & \cellcolor[rgb]{ .686,  .839,  .545}4.6 & \cellcolor[rgb]{ .663,  .831,  .541}35.5 & \cellcolor[rgb]{ .557,  .8,  .522}24.7 & \cellcolor[rgb]{ .635,  .824,  .537}49.3 & \cellcolor[rgb]{ .647,  .827,  .537}24.6 \\
    \rowcolor[rgb]{ .608,  .816,  .529} \textbf{2021} & \cellcolor[rgb]{ .745,  .859,  .561}8.3 & \cellcolor[rgb]{ .569,  .804,  .522}5.2 & \cellcolor[rgb]{ .565,  .8,  .522}39.6 & \cellcolor[rgb]{ .561,  .8,  .522}24.6 & \cellcolor[rgb]{ .573,  .804,  .522}53.0 & \cellcolor[rgb]{ .588,  .808,  .525}26.2 \\
    \rowcolor[rgb]{ .608,  .816,  .529} \textbf{2022} & \cellcolor[rgb]{ .616,  .82,  .533}10.5 & \cellcolor[rgb]{ .549,  .796,  .518}5.3 & \cellcolor[rgb]{ .506,  .784,  .51}41.9 & \cellcolor[rgb]{ .573,  .804,  .522}24.4 & \cellcolor[rgb]{ .518,  .788,  .51}56.4 & \cellcolor[rgb]{ .537,  .792,  .514}27.7 \\
    \rowcolor[rgb]{ .431,  .761,  .494} \textbf{2023} & \cellcolor[rgb]{ .525,  .788,  .514}12.1 & \cellcolor[rgb]{ .451,  .765,  .498}5.8 & \cellcolor[rgb]{ .424,  .757,  .49}45.2 & \cellcolor[rgb]{ .404,  .753,  .486}28.3 & \cellcolor[rgb]{ .404,  .753,  .486}63.0 & \cellcolor[rgb]{ .424,  .757,  .49}30.9 \\
    \rowcolor[rgb]{ .431,  .761,  .494} \textbf{2024} & \cellcolor[rgb]{ .388,  .745,  .482}14.4 & \cellcolor[rgb]{ .388,  .745,  .482}6.1 & \cellcolor[rgb]{ .388,  .745,  .482}46.6 & \cellcolor[rgb]{ .388,  .745,  .482}28.6 & \cellcolor[rgb]{ .388,  .745,  .482}63.9 & \cellcolor[rgb]{ .388,  .745,  .482}31.9 \\
    \bottomrule
    \end{tabular}%
}
  \label{tab:edu_ablations}%
\end{wraptable}

Building on \citet{gunasekar2023textbooks}, documents with higher educational values are treated as higher-quality samples—a strategy widely adopted in pre-training works.
In \finemath, we create the \finemath-Web-Pro subset from \finemath-Web data using FineMath classifier~\citep{allal2025smollm2smolgoesbig} to score documents on a 0–5 scale. 
However, we found that document distribution and relevance to mathematical reasoning vary over time. 
As Table~\ref{tab:edu_ablations} indicates, after applying Edu filtering, training 5B tokens on some years' data (e.g., 2014) yields marginal improvements, whereas later years achieve much higher performance than FineMath-4plus~(FM-4plus). Based on these observations, we adopted a dynamic filtering strategy: a more tolerant threshold (Edu score $\geq$ 3) for recent years (e.g., 2023–2024) and a stricter one (Edu score $\geq$ 4) for earlier periods (e.g., 2014–2017).
Similar to Nemontron-CC~\citep{su2024nemotron}, we also used an LLM (in our case, we use \llamaiii.3-70B-instruct) to further remove noise, and refine the web text into higher quality. Please see Figure~\ref{prompt:edu_rewriting} for the detailed prompt.
\begin{figure}[htbp]
\small
\centering
\begin{tcolorbox}
\begin{Verbatim}[breaklines=true,breaksymbol=,]
Task: 
- Carefully analyze the provided text to extract key facts, concrete details, important numbers, and core concepts. 
- Remove any irrelevant or noisy information, and reorganize the content into a logically structured, information-dense, and concise version that is easy to learn from. Output only the refined text.
- Strive to maintain the original length as much as possible (avoid excessive shortening).

Text:
<EXAMPLE>

Just output the refined text, no other text.
\end{Verbatim}
\end{tcolorbox}
\caption{Rewriting Prompt for constructing \finemath-Web-Pro.}
\label{prompt:edu_rewriting}
\end{figure}

\subsection{Further Ablation on \fasttext{}}

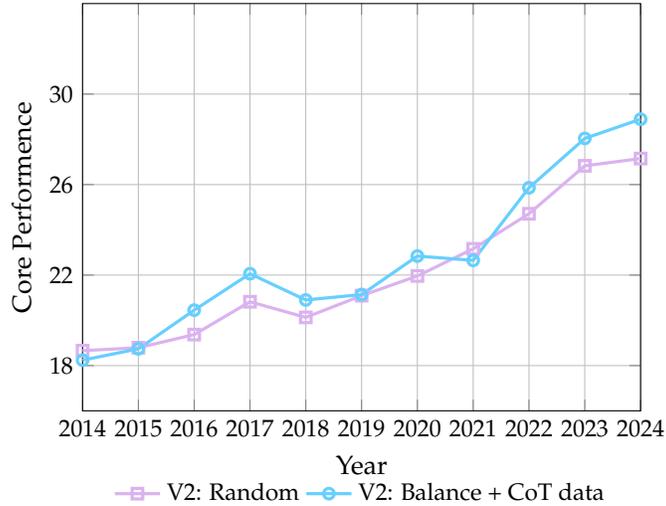
\begin{figure}[ht]
    \centering
    \begin{tikzpicture}
        \begin{axis}[
            xlabel={Year},
            ylabel={Core Performence},
            xmin=2014, xmax=2024,
            ymin=16, ymax=34,
            xtick={2014,2015,2016,2017,2018,2019,2020,2021,2022,2023,2024},
            xticklabels={2014,2015,2016,2017,2018,2019,2020,2021,2022,2023,2024},
            scaled ticks=false,
            ytick={18, 22, 26, 30},
            tick label style={font=\small},
            legend style={font=\small, draw=none, fill=none, at={(0.5,-0.15)}, anchor=north, legend columns = 2},
            grid=major,
            width=9cm,
            height=7cm
        ]

        \definecolor{colorwarmgreen}{HTML}{DBB2F0}   
        \definecolor{colorwarmred}{HTML}{66CFFE}   

        \addplot[color=colorwarmgreen, mark=square, very thick] coordinates {
            (2014, 18.66) (2015, 18.79) (2016, 19.37) (2017, 20.82) (2018, 20.13) (2019, 21.09) (2020, 21.96) (2021, 23.16) (2022, 24.71) (2023, 26.83) (2024, 27.15) };
        \addlegendentry{V2: Random}

        \addplot[color=colorwarmred, mark=o, very thick] coordinates {
(2014,18.24) (2015, 18.74) (2016, 20.45) (2017, 22.06)
(2018, 20.90) (2019, 21.14) (2020, 22.84) (2021, 22.65)
(2022, 25.86) (2023, 28.04) (2024, 28.89)
        };
        \addlegendentry{V2: Balance + CoT data}

        \end{axis}
    \end{tikzpicture}
    \caption{Ablation on \fasttext{} for each year's all dumps within 5B-token training budget}
    \label{app-fig:ablation_on_different_fasttext_all_year_appendix}
\end{figure}

We further validated our decision through experiments on all yearly dumps. Specifically, we conducted pre-training on the top 10\% highest-scoring filtered data from each yearly dump using different versions of \fasttext{}. As shown in Figure~\ref{app-fig:ablation_on_different_fasttext_all_year_appendix}, the results confirm the effectiveness of our final version (V2: Balance + CoT data), demonstrating clear improvements over the initial version in our second-round filtering. Another interesting observation is that data quality, as indicated by downstream performance, gradually improves over time.

\newpage
\section{Details for Curating \finemath-Code}
~\label{app:code}
We used Llama-3.1-70B-Instruct to annotate 25K randomly sampled code data and fine-tuned a Qwen-2.5-0.5B model to judge the code quality and decide whether to filter the code. 
The scoring prompts are presented in Figure~\ref{prompt:math_score} and Figure~\ref{prompt:edu_score}.
We only keep code data with Math Score $\mathbf{\geq 4}$ and Code Score $\mathbf{\geq 4}$, and all other code data are treated as negative samples during training.

In Table~\ref{app.tab:code_recall_tuning}, we list details for our supervised fine-tuning configurations. 
We use LlamaFactory~\citep{zheng2024llamafactory} as our code base.
Same as ProX~\citep{zhou2024programming}, we also select the model with highest F1 score as out final recalling models, which achieves 80\% on a split validation set.

\begin{table}[htbp]
  \centering
  \caption{Training parameters for SLM.}
    \begin{tabular}{l|c}
    \toprule
    \textbf{HyperParams} & \textbf{Setting} \\
    \midrule
    LR    & 1e-5 \\
    LR Schedule & cosine \\
    Batch Size & 64 \\
    Number of Epochs & 2 \\
    Context Length & 2048 \\
    \bottomrule
    \end{tabular}%
  \label{app.tab:code_recall_tuning}%
\end{table}%

\begin{figure}[htbp]
\small
\centering
\begin{tcolorbox}
\begin{Verbatim}[breaklines=true,breaksymbol=,]
Below is an extract from a resource focused on mathematical reasoning. Evaluate its educational value in effectively teaching concepts in this area, with emphasis on mathematical reasoning. Use the additive 5-point scoring system described below. Points accumulate based on each criterion:

- Add 1 point if the resource contains valid content in mathematics, reasoning, logic puzzles, or scientific computation, even if it’s not inherently educational (e.g., configurations or specialized algorithms).
- Add another point if the resource addresses practical concepts in these areas, such as solving math problems or reasoning tasks, even without annotations or explanations.
- Award a third point if the resource is suitable for educational use and introduces key concepts in mathematics or reasoning, with a structured format and some explanations or annotations.
- Give a fourth point if the resource is self-contained and directly useful for teaching, resembling a structured exercise, tutorial, or part of a lesson in mathematical reasoning or logic.
- Grant a fifth point if the resource is outstanding in educational value and perfectly suited for teaching, with clear, step-by-step explanations and thorough annotations on mathematical reasoning concepts.

The extract: <EXAMPLE>

After examining the extract:

- Briefly justify your total score, up to 100 words.
- Conclude with the score using the format: "Score: <total points>"
\end{Verbatim}
\end{tcolorbox}
\caption{Scoring Prompt for evaluating code snippets' relevance to mathematics.}
\label{prompt:math_score}
\end{figure}

\begin{figure}[!t]
\small
\centering
\begin{tcolorbox}
\begin{Verbatim}[breaklines=true,breaksymbol=,]
Below is an extract from a <CODE_TYPE> program. Evaluate whether it has a high educational value and could help teach coding. Use the additive 5-point scoring system described below. Points are accumulated based on the satisfaction of each criterion:

- Add 1 point if the program contains valid <CODE_TYPE> code, even if it's not educational, like boilerplate code, configs, and niche concepts.
- Add another point if the program addresses practical concepts, even if it lacks comments. 
- Award a third point if the program is suitable for educational use and introduces key concepts in programming, even if the topic is advanced (e.g., deep learning). The code should be well-structured and contain some comments. 
- Give a fourth point if the program is self-contained and highly relevant to teaching programming. It should be similar to a school exercise, a tutorial, or a <CODE_TYPE> course section.
- Grant a fifth point if the program is outstanding in its educational value and is perfectly suited for teaching programming. It should be well-written, easy to understand, and contain step-by-step explanations and comments.

The extract:
<EXAMPLE>

After examining the extract: 
- Briefly justify your total score, up to 100 words.
- Conclude with the score using the format: "Score: <total points>"
\end{Verbatim}
\end{tcolorbox}
\caption{Scoring Prompt for evaluating code snippets' general quality, i.e., educational value.}
\label{prompt:edu_score}
\end{figure}

\clearpage
\section{Details for Curating \finemath-Synth}

\paragraph{Synthetic Text Data} We provide the prompts for extraction and refining Q\&A below.
\begin{figure}[ht]
\small
\centering
\begin{tcolorbox}
\begin{Verbatim}[breaklines=true,breaksymbol=,]
Below is a web document extract. Assess whether it contains a mathematical question-and-answer pair:

- If the web document extract does not contain a mathematical question-and-answer pair, return the explicit symbol `[NO QA]`.
- If a mathematical question-and-answer pair is found, extract it in the following format:
  Question: <question text with complete problem statement and all necessary mathematical information>
  Answer: <complete solution with all necessary steps and calculations included> (only if an answer is provided, otherwise do not generate this line)
- The extracted pair must be self-contained and mathematically precise, allowing independent solving without additional context.

#### The extract:
<EXAMPLE>

Now process the extract and return the result.
\end{Verbatim}
\end{tcolorbox}
\caption{Prompt for QA extraction.}
\label{prompt:qa_extraction}
\end{figure}

\begin{figure}[ht]
\small
\centering
\begin{tcolorbox}
\begin{Verbatim}[breaklines=true,breaksymbol=,]
Below is a mathematical question-and-answer pair. Refine the answer based on the following requirements:

- **If the answer does not contain any explanation or intermediate reasoning process**:
  - Add only necessary intermediate reasoning process leading to the given answer
  - Ensure the added steps are logical, clear, and provide necessary explanation of the solution process

- **If the answer already includes necessary solution process**:
  - Reorganize the solution into a clear and well-structured format for better readability and understanding
  - for simple solutions, there is no need to use latex format

- Maintain the original question text and provide the refined answer in the same format:
  - Question: <question text>
  - Answer: <refined solution>

#### The question-and-answer pair:
<EXAMPLE>

Suppose you are a math teacher, you should explain the solution in a way that is easy for a student to understand. Now process the pair and return the refined result.
\end{Verbatim}
\end{tcolorbox}
\caption{Prompt for Refined QA.}
\label{prompt:qa_refinement}
\end{figure}

\newpage
\paragraph{Synthetic Code Data} We provide the prompt for code translation in Figure~\ref{prompt:code_translation} at below.
\paragraph{Synthetic Code Block Data}
\begin{figure}[htbp]
\small
\centering
\begin{tcolorbox}
\begin{Verbatim}[breaklines=true,breaksymbol=,]
Below is an extract from a code snippet. Translate the code from other programming languages into Python. Read the code carefully and translate it into Python.

- If the original code has poor quality or cannot be converted to Python, return the explicit symbol "[Untranslatable]". 
- The translated Python code should meet the following requirements:
    - Ensure good code formatting.
    - Include proper comments or explanations for clarity explaining the logic where needed.
    - Add docstrings when necessary to improve readability.
    - Wrap the generated Python code within python ```python ```.
    - Keep good test cases if any.

The extract:
```
<EXAMPLE>
```

Do not produce any additional commentary or text beyond ```python ```.
Now output the translated Python code:
\end{Verbatim}
\end{tcolorbox}
\caption{Prompt for translating non-Python code samples into Python code samples.}
\label{prompt:code_translation}
\end{figure}

We used the same prompts as in ~\citet{lu2024mathcoder2}. Please see Figure~\ref{prompt:mathcoder2}.
\begin{figure}[htbp]
\small
\centering
\begin{tcolorbox}
\begin{Verbatim}[breaklines=true,breaksymbol=,]
You will be presented with a text related to math. I need you to identify all the complex computations in it. For each complex computation that requires a scratchpad, find out the conditions needed for the computation, the latex expression that conducts the computation, and the result of the computation. Then generate a Python code snippet for each computation that demonstrates how the result is reached. Output each computation in the following format:

Conditions Needed:
1. [Condition 1]
2. [Condition 2]
...

Computation Expression:
$[Latex Expression]$

Computation Result:
[Computation Result]

Python Code Snippet:
```python
[Python Code]
```

There can be more than one complex computation in the text. Output only the computations that requires calculation. Do not include mathematical statements or definitions as a computation. Make sure each snippet can be executed individually. The text is as follows:

<EXAMPLE>

The computations are:
\end{Verbatim}
\end{tcolorbox}
\caption{Prompt for generating code-block data~\citep{lu2024mathcoder2}.}
\label{prompt:mathcoder2}
\end{figure}
Our AST filtering mainly contains the following aspects:
\begin{enumerate}
\item \textbf{Code Parsing and AST Generation}: The input code is parsed into an AST using Python's built-in \texttt{ast} module. The system first verifies code length constraints (max 100,000 characters) and handles syntax errors through exception catching.
\item \textbf{Import Declaration Analysis}: A specialized visitor collects all imported modules and their aliases through two-phase inspection:
\begin{itemize}
    \item Direct imports (\texttt{import x as y}) mapping
    \item Selective imports from modules (\texttt{from a import b as c})
\end{itemize}

\item \textbf{Semantic Node Traversal}: A secondary visitor examines all function calls and context managers, checking against three prohibition categories:
\begin{itemize}
    \item \textit{File Operations}: file I/O methods (e.g., \texttt{open}, \texttt{savefig}), path manipulations, and serialization functions
    \item \textit{Concurrency Patterns}: Thread/process creation calls and 5+ restricted modules (e.g., \texttt{threading}, \texttt{asyncio})
    \item \textit{Network Communication}: network libraries and protocol-specific methods (e.g., \texttt{requests.get}, \texttt{socket.send})
\end{itemize}

\item \textbf{Module Dependency Verification}: Cross-references imported modules against prohibited libraries spanning file systems (\texttt{shutil}), parallelism (\texttt{multiprocessing}), and network protocols (\texttt{ftplib}).

\item \textbf{Context-Specific Checks}: Special handling for:
\begin{itemize}
    \item \texttt{with} statements containing file open operations
    \item Class instantiations of thread/process primitives
    \item Path manipulation methods in object-oriented interfaces
\end{itemize}
\end{enumerate}

\section{Training Details}
\subsection{TinyLlama Training}
In all ablation experiments, we keep our training hyper-parameter the same except for training steps.
We present our full training details in Table~\ref{app-tab:training-detail}.
\begin{table}[htbp]
    \centering
    \caption{Training hyper-parameters.}
    \label{app-tab:training-detail}
    \begin{tabular}{lc}
        \toprule
        \textbf{Hyper-parameter} & \textbf{5B / 15B / 55B Tokens} \\
        \midrule
        \multirow{1}{*}{Context Length} & {2,048} \\
        \multirow{1}{*}{Batch Size} & {1,024} \\
        Max Steps & 2,500 / 7,500 / 27,500 \\
        \multirow{1}{*}{Warmup Steps} & {0} \\
        \multirow{1}{*}{Weight Decay} & {0.1} \\
        \multirow{1}{*}{Optimizer} & {AdamW} \\
        \multirow{1}{*}{LR Scheduler} & {cosine} \\
        \multirow{1}{*}{Learning Rate (LR)} & {$8\text{e-}5 \rightarrow 8\text{e-}6$} \\
        \bottomrule
    \end{tabular}
\end{table}

\begin{table}[htbp]
\centering
\caption{Training Data Mixture for \llamaiii.}
\label{app-tab:llama-training-data-mixture}
\begin{tabular}{lc}\toprule
\textbf{Data} &\textbf{Ratio \%} \\\midrule
DCLM &10 \\
Web &15 \\
Web-pro &35 \\
Code &2.5 \\
QA &10 \\
Trans. code &2.5 \\
Text \& code block &25 \\
\textbf{Total} &100 \\
\bottomrule
\end{tabular}
\end{table}

\begin{table}[htbp]
    \centering
    \caption{Training hyper-parameters.}
    \label{app-tab:llama-3training-detail}
    \begin{tabular}{lc}
        \toprule
        \textbf{Hyper-parameter} & \textbf{\llamaiii.2-1B / 3B} \\
        \midrule
        \multirow{1}{*}{Context Length} & {8,192} \\
        \multirow{1}{*}{Batch Size} & {512} \\
        Max Steps & 25,000 / 25,000~(stop at 12,500) \\
        \multirow{1}{*}{Warmup Steps} & {0} \\
        \multirow{1}{*}{Weight Decay} & {0.1} \\
        \multirow{1}{*}{Optimizer} & {AdamW} \\
        \multirow{1}{*}{LR Scheduler} & {cosine} \\
        \multirow{1}{*}{Learning Rate (LR)} & \makecell{{$5\text{e-}5 \rightarrow 5\text{e-}6$}\\{$3\text{e-}5 \rightarrow 3\text{e-}6$}} \\
        \bottomrule
    \end{tabular}
\end{table}

\subsection{\llamaiii Training}
\label{app-sec:llama-train}
The data mixture and hyper-parameters for \llamaiii training are presented in Table~\ref{app-tab:llama-training-data-mixture} and Table~\ref{app-tab:llama-3training-detail}.

\section{Evaluation Details and Full Results}
\label{app-sec:eval}

\subsection{Full Benchmarks} 

Our core set of tasks and eval settings are at below.
We revised our evaluation from DeepSeekMath~\citep{shao2024deepseekmath}: we fixed one of its prompts~\footnote{\scriptsize{\href{https://github.com/deepseek-ai/DeepSeek-Math/blob/main/evaluation/few\_shot\_prompts/pal\_math\_4\_shot.py\#L54}{https://github.com/deepseek-ai/DeepSeek-Math/blob/main/evaluation/few\_shot\_prompts/pal\_math\_4\_shot.py\#L54}}}, and support PAL for more benchmarks.

\begin{enumerate}
    \item GSM8K~\citep{cobbe2021training-gsm8k}, 8-shot
    \item MATH~\citep{hendrycks2measuring-math}, 4-shot
    \item ASDiv~\citep{miao-etal-2020-diverse-asdiv}, 8-shot 
    \item SVAMP~\citep{patel-etal-2021-nlp-svamp}, 8-shot
    \item MAWPS~\citep{koncel-kedziorski-etal-2016-mawps}, 8-shot
\end{enumerate} 

Our extended set of tasks are:
\begin{enumerate}
    \item MMLU-STEM~\citep{hendrycks2020measuring-mmlu}, 4-shot
    \item TabMWP~\citep{ludynamic-tabmwp}, 8-shot
    \item MathQA~\citep{amini-etal-2019-mathqa}, 8-shot
    \item SAT~\citep{azerbayev2023llemma}, 4-shot
    \item OCW Courses~\citep{lewkowycz2022solving-ocw}, 4-shot
\end{enumerate}

\subsection{Full Ablation Results} 
\label{app.subsec:full_ablations}

We present our full results in this section:
\begin{enumerate}
    \item For {ablation on Web Data} We provide the full ablation results on math text extraction, Minhash deduplication and \fasttext{} in Table~\ref{app-tab:full_ablation_results_on_text_extraction}, Table~\ref{app-tab:full_ablation_results_on_minhash_dedup}, Table~\ref{app-tab:full_ablation_results_on_fasttext}.
    \item For ablations on Code filtering, please see Table~\ref{app_tab:cot_pal_filter_comparison}, and Table~\ref{app_tab:cot_pal_comparison}. 
    \item For ablations on synthetic data, please see Table~\ref{app.tab:qa_ablation}, and Table~\ref{app_tab:code_synth_ablations}.
    \item The full comparison results are provided in Table~\ref{app-tab:full_comparison_results_with_existing_results}.
    \item For evaluation results for \llamaiii, please see Table~\ref{app.tab:llama3}.
\end{enumerate}

\begin{table}[htp]
\centering
\caption{Full ablation results on math text extraction within 15B-token training budget}
\resizebox{\textwidth}{!}{%
\begin{tabular}{c|c|cccccccccc|cc}
\toprule
\textbf{Text Extractors}              & \textbf{\begin{tabular}[c]{@{}c@{}} w/ HTML \\ Optimization\end{tabular}}   & \textbf{ASDiV}              & \textbf{GSM8K}             & \textbf{MATH}              & \textbf{MATH-SAT}           & \textbf{MATHQA}             & \textbf{MAWPS}              & \textbf{MMLU-STEM}          & \textbf{OCW}               & \textbf{SWAMP}              & \textbf{TABMWP}             & \textbf{Core Avg.}          & \textbf{Ext. Avg.}          \\ \midrule
{\color[HTML]{C0C0C0} TinyLlama-1.1B} & {\color[HTML]{C0C0C0} -} & {\color[HTML]{C0C0C0} 18.0} & {\color[HTML]{C0C0C0} 3.0} & {\color[HTML]{C0C0C0} 3.1} & {\color[HTML]{C0C0C0} 40.6} & {\color[HTML]{C0C0C0} 13.2} & {\color[HTML]{C0C0C0} 20.8} & {\color[HTML]{C0C0C0} 16.3} & {\color[HTML]{C0C0C0} 2.9} & {\color[HTML]{C0C0C0} 11.0} & {\color[HTML]{C0C0C0} 18.0} & {\color[HTML]{C0C0C0} 11.2} & {\color[HTML]{C0C0C0} 14.7} \\ \midrule
\trafil                                   & \ding{55}                        & 32.6                        & 5.9                        & \textbf{4.3}               & 21.9                        & 12.9                        & 44.8                        & \textbf{23.2}               & 2.2                        & \textbf{22.3}               & 21.8                        & 22.0                        & 19.2                        \\
\resili                                 & \ding{52}                       & 33.5                        & 5.8                        & 3.9                        & 15.6                        & 10.9                        & 47.3                        & 21.3                        & \textbf{2.6}               & 22.1                        & 22.7                        & 22.5                        & 18.6                        \\
\trafil                                   & \ding{52}                        & \textbf{36.3}               & \textbf{7.0}               & 3.9                        & \textbf{25.0}               & \textbf{14.7}               & \textbf{49.5}               & 22.6                        & 2.2                        & 22.1                        & \textbf{22.8}               & \textbf{23.8}               & \textbf{20.6}               \\ \bottomrule
\end{tabular}
}
\label{app-tab:full_ablation_results_on_text_extraction}
\end{table}
\begin{table}[ht]
\centering
\caption{Full ablation results on Minhash LSH within 55B-token training budget}
\resizebox{\textwidth}{!}{%
\begin{tabular}{ccc|cccccccccc|cc}
\toprule
\textbf{(r,b)}     & \textbf{t}    & \textbf{\begin{tabular}[c]{@{}c@{}} remaining \\ tokens (B)\end{tabular}} & \textbf{ASDiV} & \textbf{GSM8K} & \textbf{MATH} & \textbf{MATH-SAT} & \textbf{MATHQA} & \textbf{MAWPS} & \textbf{MMLU-STEM} & \textbf{OCW} & \textbf{SWAMP} & \textbf{TABMWP} & \textbf{Core. Avg.} & \textbf{Ext. Avg.} \\ \midrule
(14,9)           & 0.70          & 16.0                      & 26.1           & 4.9            & 3.4           & \textbf{25.0}     & 10.0            & 36.0           & 20.5               & 2.6          & 16.3           & 21.0            & 17.3                & 16.6               \\
(14,8)           & 0.75          & 23.5                      & 29.1           & \textbf{5.4}   & 3.7           & 17.5              & 9.3             & 38.6           & \textbf{23.1}      & 1.5          & \textbf{18.7}  & \textbf{23.1}   & 19.1                & 17.0               \\
\textbf{(11,10)} & \textbf{0.75} & \textbf{26.0}             & 29.8           & 4.4            & \textbf{3.9}  & 23.1              & 10.2            & \textbf{41.4}  & 19.3               & \textbf{2.9} & 17.6           & 22.1            & \textbf{19.4}       & \textbf{17.5}      \\
(11,11)          & 0.75          & 25.0                      & \textbf{30.1}  & 4.3            & 3.8           & 9.4               & 10.9            & 38.9           & 21.0               & 2.1          & \textbf{18.7}  & 20.3            & 19.2                & 16.0               \\
(9,12)           & 0.80          & 29.0                      & 28.3           & 4.4            & 3.6           & 18.8              & \textbf{11.2}   & 40.3           & 21.8               & 2.4          & 17.5           & 20.7            & 18.8                & 16.9               \\
(9,13)           & 0.80          & 30.0                      & 27.7           & 3.5            & 3.5           & 13.8              & 10.0            & 36.7           & 21.7               & 2.1          & 16.6           & 21.1            & 17.6                & 15.7               \\ \bottomrule
\end{tabular}
}
\label{app-tab:full_ablation_results_on_minhash_dedup}
\end{table}
\begin{table}[!t]
\centering
\caption{Full ablation results on \fasttext{} within 5B-token training budget}
\resizebox{\textwidth}{!}{%
\begin{tabular}{l|cccccccccc|cc}
\toprule
\textbf{\fasttext{} version} & \textbf{ASDiV} & \textbf{GSM8K} & \textbf{MATH} & \textbf{MATH-SAT} & \textbf{MATHQA} & \textbf{MAWPS} & \textbf{MMLU-STEM} & \textbf{OCW} & \textbf{SWAMP} & \textbf{TABMWP} & \textbf{Core. Avg.} & \textbf{Ext. Avg.} \\ \midrule
V1: \owm{}                        & 34.6           & 5.6            & 3.2           & 34.4              & 12.0            & 45.8           & 21.1               & 2.2          & 23.0           & 18.4            & 22.4                & 20.0               \\
V2: Random                & 41.7           & 8.6            & 5.1           & 15.6              & 11.6            & 55.9           & 17.1               & 2.2          & 24.5           & 25.1            & 27.2                & 20.7               \\
V2: Balance               & 41.3           & 8.9            & 5.0           & \textbf{28.1}     & 15.5            & 57.8           & \textbf{19.2}      & 2.2          & 26.2           & \textbf{26.2}   & 27.8                & \textbf{23.0}      \\
V2: Balance + CoT         & \textbf{44.2}  & \textbf{9.6}   & \textbf{5.4}  & 25.0              & \textbf{15.7}   & \textbf{59.0}  & 17.1               & 2.2          & \textbf{26.3}  & 25.8            & \textbf{28.9}       & \textbf{23.0}      \\ \bottomrule
\end{tabular}
}
\label{app-tab:full_ablation_results_on_fasttext}
\end{table}
\begin{table}[!t]
  \centering
  \caption{Performance comparison of CoT and PAL under different filtering criteria}
  \resizebox{0.75\textwidth}{!}{%
    \begin{tabular}{c|cccccc}
    \toprule
    \multicolumn{1}{c|}{\multirow{2}[4]{*}{\textbf{Filter Criteria}}} & \multicolumn{6}{c}{\textbf{CoT}}           \\
\cmidrule{2-7}
& \textbf{GSM8K} & \textbf{MATH} & \textbf{ASDiV} & \textbf{MAWPS} & \textbf{SVAMP} & \textbf{Avg.}\\
    \midrule
    \textbf{text only} & 4.4   & 4.1   & 29.3  & 39.5  & 17.7  & 19.0\\
    \textbf{$S\_\text{edu} \geq 3, S\_\text{math} \geq 3$} & 4.1   & 4.4   & 28.9  & 40.2  & 19.4  & 19.4\\
    \textbf{$S\_\text{edu} \geq 3, S\_\text{math} \geq 4$} & 4.9   & 4.2   & 29.8  & 41.1  & 19.2  & 19.8\\
    \textbf{$S\_\text{edu} \geq 4, S\_\text{math} \geq 3$} & 4.9   & 4.3   & 29.8  & 39.9  & 19.4  & 19.7\\
    \textbf{$S\_\text{edu} \geq 4, S\_\text{math} \geq 4$} & 4.3   & 4.2   & 29.5  & 38.5  & 17.3  & 18.8\\
    \midrule
    \multicolumn{1}{c|}{\multirow{2}[4]{*}{\textbf{Filter Criteria}}} & \multicolumn{6}{c}{\textbf{PAL}} \\
    \cmidrule{2-7}          & \textbf{GSM8K} & \textbf{MATH} & \textbf{ASDiV} & \textbf{MAWPS} & \textbf{SVAMP} & \textbf{Avg.} \\
    \midrule
    \textbf{text only} & 2.8  & 2.9   & 24.8  & 30.1  & 17.6  & 15.6 \\
    \textbf{$S\_\text{edu} \geq 3, S\_\text{math} \geq 3$} & 3.7   & 3.6   & 25.8  & 31.7  & 15.6  & 16.1 \\
    \textbf{$S\_\text{edu} \geq 3, S\_\text{math} \geq 4$} & 4.4   & 4.3   & 27.2  & 31.4  & 16.5  & 16.8 \\
    \textbf{$S\_\text{edu} \geq 4, S\_\text{math} \geq 3$} & 4.5   & 3.7   & 27.4  & 32.3  & 19.5  & 17.5 \\
    \textbf{$S\_\text{edu} \geq 4, S\_\text{math} \geq 4$} & 5.7   & 5.5   & 29.7  & 36.4  & 20.2  & 19.5 \\
    \bottomrule
    \end{tabular}%
}
  \label{app_tab:cot_pal_filter_comparison}%
\end{table}%
\begin{table}[!t]
  \centering
  \caption{Performance comparison of CoT and PAL under different mix ratios.}
  \resizebox{0.75\textwidth}{!}{%
    \begin{tabular}{c|cccccc}
    \toprule
    \multicolumn{1}{c|}{\multirow{2}[4]{*}{\textbf{Mix Ratio}}} & \multicolumn{6}{c}{\textbf{CoT}}             \\
\cmidrule{2-7}          & \textbf{GSM8K} & \textbf{MATH} & \textbf{ASDiV} & \textbf{MAWPS} & \textbf{SVAMP} & \textbf{Avg.} \\
    \midrule
    \textbf{text only} & 4.4   & 4.1   & 29.3  & 39.5  & 17.7  & 19.0   \\
    \textbf{code:text = 1:7} & 3.8   & 4.2   & 30.2  & 40.0  & 18.3  & 19.3   \\
    \textbf{code:text = 1:4} & 4.6   & 4.1   & 29.5  & 40.6  & 18.9  & 19.5   \\
    \textbf{code:text = 1:2} & 3.9   & 4.0   & 28.3  & 27.6  & 18.1  & 16.4   \\
    \textbf{code:text = 1:1} & 3.6   & 3.9   & 26.7  & 36.7  & 16.6  & 17.5   \\
    \midrule
    \multicolumn{1}{c|}{\multirow{2}[4]{*}{\textbf{Mix Ratio}}} & \multicolumn{6}{c}{\textbf{PAL}} \\
    \cmidrule{2-7}          & \textbf{GSM8K} & \textbf{MATH} & \textbf{ASDiV} & \textbf{MAWPS} & \textbf{SVAMP} & \textbf{Avg.} \\
    \midrule
    \textbf{text only} & 2.8  & 2.9   & 24.8  & 30.1  & 17.6  & 15.6 \\
    \textbf{code:text = 1:7} & 4.4   & 4.3   & 27.2  & 31.8  & 19.5  & 17.4 \\
    \textbf{code:text = 1:4} & 4.4   & 4.4   & 29.2  & 33.6  & 20.2  & 18.4 \\
    \textbf{code:text = 1:2} & 4.3   & 4.4   & 27.6  & 34.1  & 17.1  & 17.5 \\
    \textbf{code:text = 1:1} & 5.4   & 4.9   & 29.5  & 36.3  & 17.7  & 18.8 \\
    \bottomrule
    \end{tabular}%
}
  \label{app_tab:cot_pal_comparison}%
\end{table}%
\begin{table}[htbp]
  \centering
  \caption{Performance comparison of CoT using different Q\&A datasets}
  \resizebox{0.75\textwidth}{!}{%
    \begin{tabular}{l|cccccc}
    \toprule
    \textbf{Data} & \textbf{ASDiV} & \textbf{GSM8K} & \textbf{MATH} & \textbf{MATH-SAT} & \textbf{MATHQA} & \textbf{MAWPS} \\
    \midrule
    \textbf{FM-4plus} & 41.9  & 10.5  & 6.1   & 34.4  & 14.4  & 57.9 \\
    \textbf{WebInstruct} & 49.5  & 13.1  & 10.6  & 25.0  & 14.7  & 65.8 \\
    \textbf{Vanilla Prompt} & 57.4  & 22.1  & 10.5  & 25.0    & 16.5  & 68.6 \\
    \textbf{ w. ELI5} & 58.6  & 25.9  & 12.3  & 21.9  & 18.4  & 71.6 \\
    \textbf{w. ELI5 + IC} & 68.0    & 33.3  & 15.3  & 34.4  & 21.7  & 79.6 \\
    \midrule
    \textbf{Data} & \textbf{MMLU-STEM} & \textbf{OCW} & \textbf{SVAMP} & {\textbf{TABMWP}} & \textbf{Core Avg.} & \textbf{Ext. Avg.} \\
    \midrule
    \textbf{FM-4plus} & 20.6  & 2.9   & 25.3  & 25.5  & 28.3  & 19.6 \\
    \textbf{WebInstruct} & 16.0    & 3.3   & 34.0    & 29.2  & 34.6  & 17.6 \\
    \textbf{Vanilla Prompt} & 17.7  & 2.9   & 37.2  & 35.4  & 39.2  & 19.5 \\
    \textbf{ w. ELI5} & 15.6  & 4.0     & 38.1  & 35.9  & 41.3  & 19.2 \\
    \textbf{w. ELI5 + IC} & 18.3  & 3.3   & 48.0    & 40.1  & 48.8 & 23.6 \\
    \bottomrule
    \end{tabular}%
    }
  \label{app.tab:qa_ablation}%
\end{table}%

\begin{table}[!t]
  \centering
  \caption{Performance comparison of CoT and PAL under different mix ratios.}
  \resizebox{0.75\textwidth}{!}{%
    \begin{tabular}{l|cccccc}
    \toprule
    \multicolumn{1}{c|}{\multirow{2}[4]{*}{\textbf{Data}}} & \multicolumn{6}{c}{\textbf{CoT}}             \\
\cmidrule{2-7}          & \textbf{GSM8K} & \textbf{MATH} & \textbf{ASDiV} & \textbf{MAWPS} & \textbf{SVAMP} & \textbf{Avg.} \\
    \midrule
    \textbf{code} & 4.3   & 4.2   & 29.5  & 38.5  & 17.3  & 18.8   \\
    \textbf{trans. code} & 3.5   & 4.3   & 30.2  & 39.3  & 17.8  & 19.0   \\
    \textbf{text \& code block} & 6.7   & 5.2   & 34.0  & 45.4  & 21.2 & 22.5   \\
    \textbf{text \& code block} (full) & 12.4   & 7.9   & 43.8  & 58.6  & 31.2  & 30.8   \\
    \midrule
    \multicolumn{1}{c|}{\multirow{2}[4]{*}{\textbf{Data}}} & \multicolumn{6}{c}{\textbf{PAL}} \\
    \cmidrule{2-7}          & \textbf{GSM8K} & \textbf{MATH} & \textbf{ASDiV} & \textbf{MAWPS} & \textbf{SVAMP} & \textbf{Avg.} \\
    \midrule
    \textbf{code} & 5.7  & 5.5   & 29.7  & 36.4  & 20.2  & 19.5 \\
    \textbf{trans. code} & 7.0   & 5.3   & 31.3  & 39.4  & 20.1  & 20.6   \\
    \textbf{text \& code block} & 9.6   & 10.6   & 41.1  & 51.2  & 27.8  & 28.1   \\
    \textbf{text \& code block} (full) & 26.9   & 17.3   & 62.1  & 78.0  & 48.3  & 46.5   \\
    \bottomrule
    \end{tabular}%
}
  \label{app_tab:code_synth_ablations}%
\end{table}%
\begin{table}[ht]
\centering
\caption{Full comparison CoT results with existing corpora
within 55B-token training budget}
\resizebox{\textwidth}{!}{%
\begin{tabular}{lcccccccccc|cc}
\toprule
\textbf{Corpus}                          & \textbf{ASDiV} & \textbf{GSM8K} & \textbf{MATH} & \textbf{MATH-SAT} & \textbf{MATHQA} & \textbf{MAWPS} & \textbf{MMLU-STEM} & \textbf{OCW} & \textbf{SWAMP} & \textbf{TABMWP} & \textbf{Core. Avg.} & \textbf{Ext. Avg.} \\ \midrule
\textbf{MegaMath-Web-Pro (15B, Ours)}    & \textbf{61.9}  & \textbf{24.1}  & \textbf{12.0} & 34.4              & 15.4            & \textbf{75.7}  & \textbf{28.2}      & 2.6          & \textbf{42.9}  & 32.5            & \textbf{43.3}       & \textbf{33.0}      \\
 FineMath-4+ (11B)                       & 55.7           & 21.1           & 11.4          & 31.3              & \textbf{23.7}   & 70.9           & 25.9               & 2.6          & 35.9           & 32.6            & 39.0                & 31.1               \\
\textbf{MegaMath-Web-Top 50\% (Ours) } & 53.0           & 15.5           & 8.4           & 31.3              & 15.5            & 68.3           & 25.7               & 3.7          & 33.1           & \textbf{32.7}   & 35.6                & 28.7               \\
 FineMath-3+ (41.6B)                     & 50.8           & 17.1           & 8.5           & 21.9              & 17.5            & 68.4           & 24.4               & \textbf{4.4} & 30.7           & 30.5            & 35.1                & 27.4               \\
\textbf{MegaMath-Web-Top 75\% (Ours) }   & 46.8           & 11.7           & 6.9           & \textbf{43.8}     & 17.3            & 62.4           & 17.2               & 2.6          & 29.2           & 30.0            & 31.4                & 26.8               \\
 InfiMM-WebMath (55B)                    & 46.1           & 12.3           & 6.4           & 25.0              & 15.3            & 63.0           & 22.5               & 3.3          & 26.3           & 28.4            & 30.8                & 24.9               \\
\textbf{MegaMath-Web-Full (Ours) } & 44.7           & 11.6           & 6.4           & 25.0              & 13.0            & 61.0           & 21.8               & 2.2          & 26.2           & 30.0            & 30.0                & 24.2               \\
 Open-Web-Math (14.5B)                   & 39.7           & 8.7            & 6.3           & 31.3              & 12.9            & 54.2           & 22.7               & 2.6          & 24.1           & 25.1            & 26.6                & 22.8               \\ \bottomrule
\end{tabular}
\label{app-tab:full_comparison_results_with_existing_results}
}
\end{table}
\begin{table}[htbp]
\centering
\caption{Full results of training \finemath on \llamaiii series of models.}
\label{app.tab:llama3}
\scriptsize
\begin{tabular}{l|cccccc}\toprule
\multirow{2}{*}{\textbf{Model}} &\multicolumn{6}{c}{\textbf{CoT}} \\
\cmidrule{2-7}
&\textbf{ASDiV} &\textbf{GSM8K} &\textbf{MATH} &\textbf{MAWPS} &\textbf{SVAMP} &\textbf{Avg.} \\
\midrule
\textbf{Llama-3.2-1B} &33.8 &8.5 &4.6 &43.3 &21.5 &22.3 \\
\textbf{MegaMath Llama-3.2-1B} &59.8 &25.7 &9.5 &74.6 &41.3 &42.2 \\
\textbf{Llama-3.2-3B} &60.5 &30.1 &9.2 &80.5 &52.6 &46.6 \\
\textbf{MegaMath Llama-3.2-3B} &78.8 &56.2 &25.1 &90.2 &71.6 &64.4 \\
\midrule
\multirow{2}{*}{\textbf{Model}} &\multicolumn{6}{c}{\textbf{PAL}} \\
\cmidrule{2-7}
&\textbf{ASDiV} &\textbf{GSM8K} &\textbf{MATH} &\textbf{MAWPS} &\textbf{SVAMP} &\textbf{Avg.} \\
\midrule
\textbf{Llama-3.2-1B} &13.4 &7.9 &3.1 &16.8 &8.4 &9.9 \\
\textbf{MegaMath Llama-3.2-1B} &42.8 &16.8 &6.3 &52.7 &29.8 &29.7 \\
\textbf{Llama-3.2-3B} &65.1 &35.7 &0.4 &83.3 &58.3 &48.6 \\
\textbf{MegaMath Llama-3.2-3B} &78.1 &55.7 &24.6 &93.7 &74.4 &65.3 \\
\bottomrule
\end{tabular}
\end{table}

\end{document}